\documentclass[10pt,twocolumn,letterpaper]{article}

\usepackage[pagenumbers]{cvpr} 

\usepackage{xcolor}
\input{preamble}

\definecolor{cvprblue}{rgb}{0.21,0.49,0.74}
\usepackage[pagebackref,breaklinks,colorlinks,citecolor=cvprblue]{hyperref}

\def\paperID{2555} 
\def\confName{CVPR}
\def\confYear{2024}

\title{CurveCloudNet: Processing Point Clouds with 1D Structure}

\author{
Colton Stearns\\
Stanford University\\
\and
Davis Rempe\\
Stanford University\\
\and
Alex Fu\\
Stanford University\\
\and
Jiateng Liu\\
Zhejiang University\\
\and
Sébastien Mascha\\
Summer Robotics\\
\and
Jeong Joon Park\\
Stanford University\\
\and
Despoina Paschalidou\\
Stanford University\\
\and
Leonidas J. Guibas\\
Stanford University\\
}

\begin{document}
\twocolumn[{%
\renewcommand\twocolumn[1][]{#1}%
\maketitle
\begin{center}
    \centering
    \vspace{-5mm}
    \captionsetup{type=figure}
    \includegraphics[width=1.0\textwidth,height=4.5cm]{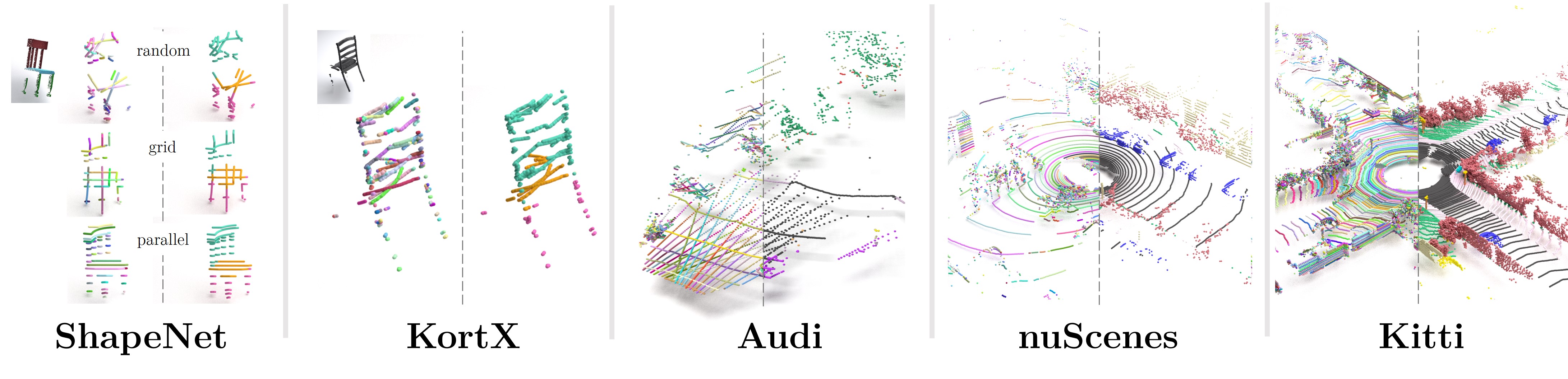}
    \vspace{-5mm}
    \captionof{figure}{Visualizations of the input curve cloud (left) and CurveCloudNet's segmentation prediction (right) for each of our five evaluation datasets. Each evaluation dataset exhibits distinct size, structure, and laser scanning pattern, as shown in \cref{tab:summary-results-and-datasets}}
    \label{fig:teaser}
\end{center}%
}]


\begin{abstract}

Modern depth sensors such as LiDAR operate by sweeping laser-beams across the scene, resulting in a point cloud with notable 1D curve-like structures. In this work, we introduce a new point cloud processing scheme and backbone, called \arch, which takes advantage of the curve-like structure inherent to these sensors. While existing backbones discard the rich 1D traversal patterns and rely on generic 3D operations, \arch parameterizes the point cloud as a collection of polylines (dubbed a ``curve cloud”), establishing a local surface-aware ordering on the points. By reasoning along curves, \arch captures lightweight curve-aware priors to efficiently and accurately reason in several \textbf{diverse} 3D environments. 
We evaluate \arch on multiple synthetic and real datasets that exhibit distinct 3D size and structure.
We demonstrate that \arch outperforms both point-based and sparse-voxel backbones in various segmentation settings, notably scaling to large scenes better than point-based alternatives while exhibiting improved single-object performance over sparse-voxel alternatives.
In all, \arch is an efficient and accurate backbone that can handle a larger variety of 3D environments than past works. 
\vspace{-1em}
\end{abstract}

\section{Introduction}
\label{sec:intro}
\begin{table*}[t]
\centering
    \begin{subtable}[t]{0.45\textwidth}
    \centering
    \scalebox{0.75}{
    \setlength{\tabcolsep}{4pt}
    \begin{tabular}{ l l | c | c c c c c}
    \toprule
      \textit{\textbf{Method / mIOU ($\uparrow$)}} & \textit{Type} & \textit{AVG} & \rotbox{0}{KortX} & \rotbox{0}{ShapeNet} & \rotbox{0}{A2D2} & \rotbox{0}{nuScenes} & \rotbox{0}{Kitti}\\
      \midrule
      PointNet++~\cite{Qi2017NIPS} & Point & 62.2 & 71.0 & 80.1 & 46.5 & 51.1 & --\\
      CurveNet~\cite{Xiang2021ICCV} & & 52.9 & 71.5 & \underline{82.8} & 4.4 & -- & -- \\
      PointMLP~\cite{Ma2022ICLR} & & 62.3 & \underline{75.4} & 80.9 & 47.6 & 67.9 & 39.5 \\ 
      PointNext~\cite{Qian2022PointNeXtRP} & & 66.6 & 73.7 & \underline{82.8} & 45.0 & 65.0 & -- \\ \hline
      MinkowskiNet~\cite{Choy2019CVPR} & Voxel & 67.6 & 60.1 & 81.1 & 53.8 & 76.2 & 66.8 \\
      Cylinder3D~\cite{Zhou2020ARXIV} & & 67.6 & 63.5 & 79.6 & 53.0 & 76.1 & 65.9\\
      SphereFormer~\cite{lai2023spherical} & & 70.6 & 69.7 & 79.5 & \textbf{55.1} & \textbf{79.5} & \underline{69.0}\\
      \hline
      CurveCloudNet (ours) & Curve & \textbf{72.7} & \textbf{78.9} & \textbf{83.1} & \underline{54.1} & \underline{78.0} & \textbf{69.5} \\
      \bottomrule
    \end{tabular}
    }
    \label{tab:summary-results}
    \end{subtable}
    \hfill
    \begin{subtable}[h]{0.45\textwidth}
    \centering
    \scalebox{0.75}{
    \setlength{\tabcolsep}{4pt}
    \begin{tabular}{ l | c c c c c }
      \toprule
      \textit{\textbf{Dataset Statistic}} & ShapeNet & KortX & A2D2 & nuScenes & Kitti \\
      \midrule
      location & Synthetic & Indoor & Outdoor & Outdoor & Outdoor \\
      scale & $\pm$ 1 & $\pm$ 2m & $+$ 70m  & $\pm$ 50m & $\pm$ 50m\\
      laser pattern & ALL & Random & Grid & Parallel & Parallel \\
      \# points & 2048 & 2048 & $\sim$ 8K & $\sim$ 35K & $\sim$ 100K \\
      \# train & 12K & 6K & 18K & 28K & 19K\\
      train/val gap & \xmark & \cmark & \xmark & \xmark & \xmark \\
      \bottomrule
    \end{tabular}
    }
    \label{tab:dataset-overview}
    \end{subtable}
    \caption{\textit{Dataset and Performance Overview} \textbf{(Left)} CurveCloudNet achieves the best mIOU on average and is best or second-best for every dataset. Empty entries indicate excessive training time that exceeds 20 days. Validation splits are reported because not all baselines were submitted to test servers. \textbf{(Right)} We evaluate on five segmentation benchmarks that exhibit diverse size, structure, and training settings. Refer to \cref{fig:teaser} for illustrations of the parallel, random, and grid laser patterns.}
    \label{tab:summary-results-and-datasets}
    \vspace{-4mm}
\end{table*}

Over the past decade, the computer vision community has proposed many backbones for processing 3D point clouds for fundamental tasks such as semantic segmentation \cite{Qi2017CVPR, Qi2017NIPS, Wang2019SIGGRAPHb, Thomas2019ICCV, Hu2020CVPR}
and object detection \cite{Wu2022GeosciRemoteSens, Wang2021NeurIPS, Wu2022ARXIV, Wu2022CVPR, Yang2022ECCV}.
Existing 3D backbones can be generally characterized as point-based or discretization-based.
Backbones that directly operate on 3D points \cite{Qi2017NIPS, Su2018CVPR, Hua2018CVPR, Wang2018CVPRb, Xu2018ECCV, Esteves2018ECCV,
Wu2019CVPR, Thomas2019ICCV} typically exchange and aggregate point features in Euclidean space, and have shown success for individual objects or relatively small indoor scenes. These methods, however, do not scale well to large scenes (\eg in outdoor settings) due to inefficiencies in processing large unstructured point sets.
On the other hand, popular discretization approaches such as sparse voxel methods \cite{Wu2018ICRA, Su2018CVPR, Graham2018CVPR, Choy2019CVPR,
Liu2019NeurIPS, Zhou2020ARXIV, Hu2020CVPR, Zhang2022ARXIV} rely on efficient sparse data structures that scale better to large scenes. However, for small or irregularly-distributed point sets, they often incur discretization errors. 

In recent years, this trade off between point and voxel backbones has been less explored due to the distinct environments in most 3D applications - autonomous vehicles do not leave roads, manufacturing robots do not leave warehouses, and quality-assurance systems do not look beyond a tabletop. However, as the community moves to dynamic and unregulated settings such as open-world robotics (\eg embodied agents), it is essential to have architectures that consistently perform well in diverse settings.

To this end, we present a novel point cloud processing scheme that achieves both performance and flexibility across diverse 3D environments. We achieve this by tailoring 
%
our approach to the popular family of laser-scanning 3D sensors (such as LiDARs), which gather 3D measurements by sweeping laser-beams across the scene.
While previous works ignore the innate curve-like structures of the scanner outputs, we parameterize the point cloud as a collection of polylines, which we refer to as a ``curve cloud”. Our formulation establishes a local structure on the points, allowing for efficient and cache-local communication between points along a curve. This enables scaling to large scenes without incurring discretization errors and/or computational overhead. Furthermore, we hypothesize that the local curve ordering injects a lightweight and flexible surface-aware prior into the network (see \cref{subsec:why-curves}).


We propose a new backbone, \arch, that applies 1D operations along curves and combines curve operations with state-of-the-art point-based operations \cite{Qian2022PointNeXtRP, Ma2022ICLR, Wang2019SIGGRAPHb, Qi2017NIPS}. \arch uses curve operations at higher resolutions when there is clearer curve structure and uses point operations at downsampled resolutions. 
Put together, \arch is an efficient, scalable, and accurate backbone that can outperform segmentation and classification pipelines in a variety of settings (see \cref{tab:summary-results-and-datasets}a).

We evaluate CurveCloudNet on a variety of object-level and outdoor scene-level datasets that exhibit distinct 3D size, structure, and unique laser scanning patterns (see \cref{tab:summary-results-and-datasets}b and \cref{fig:teaser}): this includes indoor, outdoor, object-centric, scene-centric, sparsely scanned, and densely scanned scenes. We evaluate \arch on the object part segmentation task using the ShapeNet \cite{Chang2015ARXIV, Yi2016ToG} dataset along with a new real-world object-level dataset captured with the \kortx scanning system \cite{summer-robotics}. For the outdoor semantic segmentation task, we use the nuScenes~\cite{Caesar2020CVPR}, Audi Autonomous Driving (A2D2)~\cite{Geyer2020ARXIV}, and Semantic Kitti ~\cite{behley2019iccv, Geiger2012CVPR} datasets.
Supplementary experiments on object classification demonstrate flexibility to other perception tasks. Our evaluations demonstrate that using curve structures leads to improved or competitive performance on \textit{all} experiments, with the best performance on average (see \cref{tab:summary-results-and-datasets}a).

In summary, we make the following \textbf{contributions}: 
\textbf{(1)} we propose operating on laser-scanned point cloud data using a \emph{curve cloud} representation, \textbf{(2)} we design efficient operations that run on polyline curves, \textbf{(3)} we design a novel backbone, CurveCloudNet, that strategically combines both curve and point operations, and \textbf{(4)} we show accurate and efficient segmentation results on real-world data captured for both objects and large-scale scenes in multiple environments and with various scanning patterns.

\begin{figure*}
    \centering
    \includegraphics[width= 0.92\textwidth]{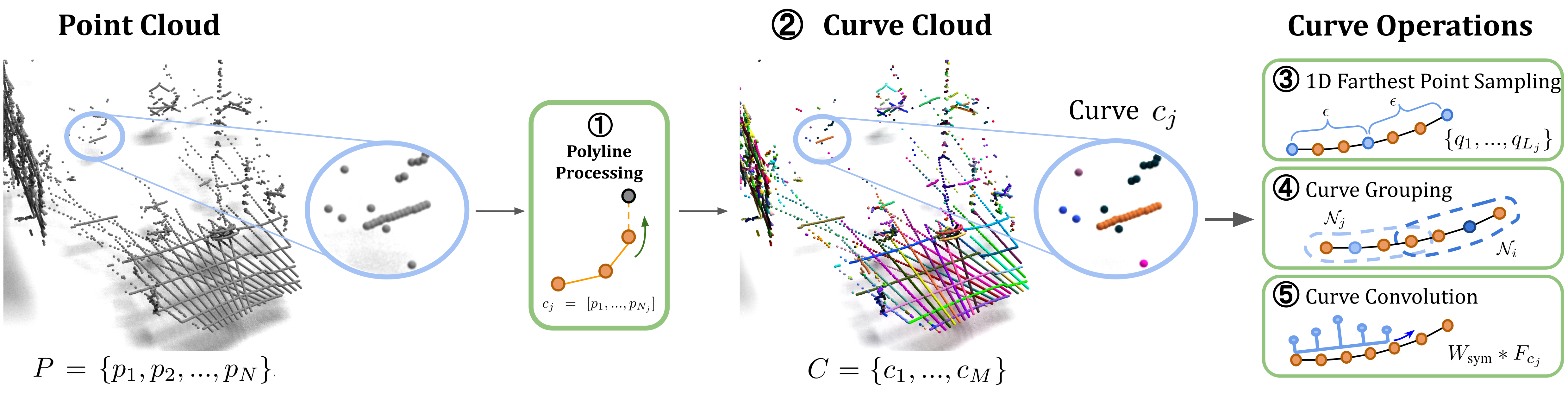}
    \caption{\textit{Overview of Curve Cloud Reasoning.} Starting from laser-scanned input data, we \textcircled{1} link points into polylines to \textcircled{2} parameterize the point cloud as a curve cloud (see \cref{sec:prelims}). We develop operations for learned architectures to specifically exploit the curve structure, including \textcircled{3} 1D farthest-point-sampling along a curve, \textcircled{4} curve grouping, and \textcircled{5} symmetric curve convolutions (see \cref{sec:curve-ops}).}
    \label{fig:overview}
    \vspace{-3mm}
\end{figure*}
\section{Related Work}

Existing point cloud methods can be roughly characterized as
point-based and discretization-based approaches. As our work addresses trade-offs between them, we discuss related works from each category.

\paragraph{Point-Based Networks}
%
One popular approach to point-based reasoning is to aggregate local neighborhood information in a hierarchical manner and at multiple geometric scales \cite{Ma2022ICLR, Qi2017NIPS, Li2018CVPR, Zhao2021ICCV,Qian2022PointNeXtRP, Hu2020CVPR, Zhang2022CVPR, Yang20203DSSDP3, Qi2017CVPR}. Recently, Ma \etal \cite{Ma2022ICLR} introduced a modern MLP-based architecture along these lines, while Quian \etal \cite{Qian2022PointNeXtRP} modernized the seminal PointNet++ \cite{Qi2017NIPS} -- both showed compelling results on object-level and indoor scenes.
Nevertheless, most hierarchical and MLP point networks are inefficient in large-scale settings, and although several backbones \cite{Hu2020CVPR,Zhang2022CVPR,Yang20203DSSDP3} have addressed this, they trade off scalability with task-specific frameworks or lower accuracy. 
In contrast, CurveCloudNet scales to large scenes by using the explicit curve structure of laser scanners.

Another line of research makes use of point convolutions for learning per-point local features. Point convolutions are usually defined using \textit{kernels} \cite{Su2018CVPR, Hua2018CVPR, Wang2018CVPRb, Xu2018ECCV, Esteves2018ECCV,
Wu2019CVPR, Mao2019ICCV, Lei2019CVPR, Komarichev2019CVPR, Lan2019CVPR, Thomas2019ICCV, Wiersma2022SIGGRAPH} or \textit{graphs} \cite{Te2018SIGGRAPH, Liu2019ICCVb, Chen2020CVPRa, Eliasof2020NeurIPS}. Kernels have been defined using a family of polynomial
functions~\cite{Xu2018ECCV}, using MLPs~\cite{Wang2018CVPRb,
Liu2019CVPR}, or directly using local 3D point coordinates \cite{Atzmon2018SIGGRAPH, Thomas2019ICCV, Wu2019CVPR, Xu2021CVPR,
Boulch2020ACCV}. In contrast, graph methods usually construct a K-Nearest-Neighbors graph in Euclidean space \cite{Shen2018CVPR, Verma2018CVPR} or feature space \cite{Wang2019SIGGRAPHb}, and apply graph-convolutions on the resulting edges and vertices.
More recently, CurveNet~\cite{Xiang2021ICCV} applied guided random walks on uniformly-sampled input point clouds to construct graph neighborhoods that go beyond K-Nearest-Neighbors and that exhibit 1D ``curve" structure; then, CurveNet pooled features over the traversed curves. Aside from defining ``curves", CurveNet and CurveCloudNet have little in common: CurveNet's guided random walks are not related to physical laser traversals and do not scale to large scenes. In contrast, \arch efficiently recovers explicit curves from a scanner’s physical laser traversals, and then applies a variety of operations, \eg, subsampling, aggregation, and convolution, along each curve.



Many works \cite{Xie2018CVPR, Liu2019AAI, Yang2019CVPR, Zhang2019CVPR, Hehe2021CVPR, Zhao2021ICCV, Ishan2021ICCV, wu2022pointtransformerv2} have shown success with \emph{attention-based aggregation} using
transformer architectures with self-attention \cite{Vaswani2017NIPS}. However, we found CurveCloudNet's reasoning over local 1D ``curve" neighborhoods to be sufficiently expressive without attention.

\paragraph{Discretization-Based Networks}
Although point-based backbones can successfully process individual objects or
small indoor scenes, they struggle to scale to large point clouds due to inefficiency in processing large unstructured point sets. To address this,
several works \cite{Su2018CVPR, Graham2018CVPR, Choy2019CVPR,
Liu2019NeurIPS, Zhang2020CVPR, Zhou2020ARXIV, Zhang2022ARXIV, Yan2018Sensors, Lang2018CVPR, Liu2021TPAMI, Xu2021ICCV, Hou2022CVPR} proposed to convert a point cloud into a 3D
voxel grid and use this volumetric representation. Early works converted a point cloud into a dense voxel grid and applied dense 3D convolutions \cite{Maturana2015IROS, Qi2016VolumetricAMCVPR}, however the cubic size of the dense grid proved to be computationally prohibitive. 
%
To scale to large scenes, several works \cite{Choy2019CVPR, Lang2018CVPR, Zhou2020ARXIV, Liu2021TPAMI, Xu2021ICCV, Cheng2021AF2S3NetAF, Xu2021RPVNetAD, Yan2020SparseSS, Hou2022CVPR, Yan20222DPASS2P} employed the sparse-voxel data structure from \cite{Graham2018CVPR}. 
MinkowskiNet~\cite{Choy2019CVPR} was a seminal work in showing that sparse voxel convolutions can be highly efficient and expressive. 
PVNAS~\cite{Liu2021TPAMI} incorporated a network architecture search, demonstrating the importance of the architecture channels, network depth, kernel sizes, and training schedule. More recent works have supplemented sparse-voxel backbones with attention operations \cite{Cheng2021AF2S3NetAF}, range-view and point information \cite{Xu2021RPVNetAD}, image information \cite{Yan20222DPASS2P}, and knowledge distillation \cite{Hou2022CVPR}. 

Other methods seek a better discretization of point clouds captured with \lidar scans. For example, PolarNet~\cite{Zhang2020CVPR}
proposed to partition input points using grid cells defined in a polar
coordinate system, while Cylinder3D~\cite{Zhou2020ARXIV} employed a cylindrical
partitioning scheme based on a cylindrical coordinate system. Sphereformer \cite{lai2023spherical} combined polar grid cells with modern attention operations. In an alternative line of research, many methods \cite{Kong2023Preprint, Gu2022MaskRangeAM, Cheng2022CenetTC, Ando2023RangeViTTV, Zhao2021FIDNetLP, Xu2020ECCV, Wu2018ICRA, Wu2019CVPR}
employ spherical or bird's-eye view projections to represent point clouds as images that are passed to a 2D convolutional or transformer network. 

Unlike discretization methods, \arch directly operates on points and curves, scaling to larger scenes without discretizing. Additionally, our curve operations are applied locally and do not assume global patterns such as polar, cylindrical, or planar structure. 


\section{Method: Learning on Curve Clouds}
Our method takes as input a 3D point cloud, parses it into a curve cloud representation, and then processes the resulting curves by leveraging specialized curve operations, as shown in \cref{fig:overview}. We focus on object part segmentation, semantic scene segmentation, and object classification, although in principle, our method is suitable for more perception tasks. 

\begin{figure*}
    \centering
    \includegraphics[width= 0.82\textwidth]{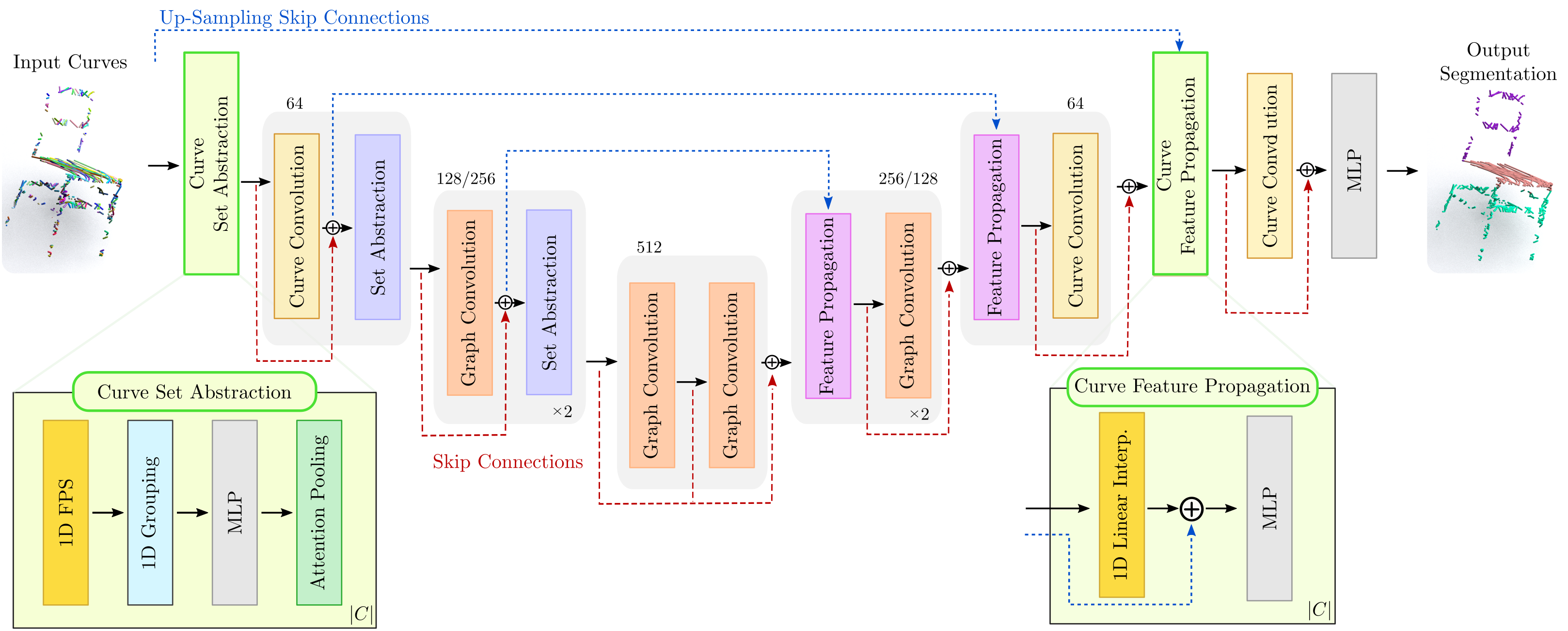}
    \caption{\textit{CurveCloudNet Architecture.} The network employs a mix of curve and point layers to process a curve cloud through progressive down-sampling followed by up-sampling with skip connections. Curve layers operate on higher resolutions to efficiently capture the 1D structure, while at lower resolutions point layers propagate information across curves. Feature dimensions are listed above each block.}
    \label{fig:curvecloudnet-architecture}
    \vspace{-2mm}
\end{figure*}

\subsection{Constructing Curve Clouds} \label{sec:prelims}
\paragraph{Problem Formulation}
The input to our model is the output of a laser-based 3D sensor represented as a point cloud $P = \{p_1, p_2, ..., p_N\}$, where $p_i = [x_i, y_i, z_i]$ is the 3D coordinates of the $i$-th point.
For each point, we are also given an associated acquisition timestamp $t_i$ and an integer laser beam ID $b_i \in [1, B]$, which are readily available from sensors like \lidar. 
For a scanner with $B$ unique laser beams, $b_i$ indicates which beam captured the point while $t_i$ gives the ordering in which points were captured.
Timestamps differ only by microseconds and indicate point ordering for constructing the curve cloud; otherwise, the point cloud is treated as an instantaneous capture of the scene.

We assume that each laser beam in the scanner captures 3D points \textit{sequentially} and at a \textit{high sampling rate} as it sweeps the scene.
Concretely, if points $p_1$, $p_2$, and $p_3$ are recorded consecutively by beam $b$, then their timestamps are ordered $t_1{<}t_2{<}t_3$. 
Moreover, if two consecutive points are spatially farther apart than some small threshold $\delta$, we assume there is a surface discontinuity and the points lie on different surfaces in the scene.

\paragraph{Curve Clouds}
A curve $c_j = [p_1, ..., p_{N_j}]$ is defined as a sequence of $N_j$ points where consecutive point pairs are connected by a line segment, \ie, a \textit{polyline}. 
The curve is bi-directional and is equivalently defined as $c_j = [p_{N_j}, p_{N_j-1}, ..., p_1]$.
A \textit{curve cloud} $C = \{c_1, ..., c_M\}$ is an unordered set of $M$ curves where each curve may contain a different number of points.
In practice, we can store a curve cloud $C$ as an $N{\times}3$ tensor of points, with an additional $ptr$ tensor of length $M$ that specifies the indices where one curve ends and the next begins.
Converting the input point cloud $P$ to a corresponding curve cloud is straightforward and extremely efficient: points from each beam $b_i$ are ordered by  timestamp and split into curves based on distances between consecutive points.
If the distance between two consecutive points is more than a set threshold $\delta$ while traversing the points in time order, then the current curve ends and a new curve begins, \ie, a surface discontinuity has occurred.
In practice, we parallelize this process across all points and laser beams on the GPU.
More details regarding the conversion process are provided in the supplement.

\noindent \textbf{Why Use Curves?} \label{subsec:why-curves}
Curve clouds inherit the benefits associated with the 3D point cloud representation including lightweight data structures and no need to discretize the space. 
But operating on curve clouds also has several advantages over point clouds.
Point clouds are highly unstructured, making operations like nearest neighbor queries and convolutions expensive. 
Curve clouds add structure through point ordering along the polylines, allowing curves to be treated as 1D grids that permit greater cache-locality, constant-time neighborhood queries, and efficient convolutions.
This structure is flexible to any laser scanning pattern unlike, \eg, cylindrical voxel grids and polar range-view projections. 
In principle, the curve structure should also bring out the geometric properties of the surface it represents, such as curvature, tangents, and boundaries.

\subsection{Operating on Curves} \label{sec:curve-ops}
We now discuss the fundamental operations for curves. 
We first introduce the network layers for curves followed by the details of curve operations used in these layers.

\vspace{-2mm}
\subsubsection{Curve Layers} \label{subsec:curve-layers}
\vspace{-1mm}

\paragraph{Curve Set Abstraction (Curve SA)}
Inspired by set abstraction layers from prior point-based work~\cite{Qi2017NIPS}, this layer adopts a curve-centric procedure to downsample the number of points on the curves. For each curve, Curve SA (1) samples a subset of ``centroid'' points along the curve using our \textit{1D farthest point sampling} algorithm, (2) groups points around these centroids in local neighborhoods using \textit{curve grouping}, (3) translates points into the local frame of their centroid and processes them with a shared MLP, and (4) pools over each local neighborhood to get a downsampled curve with associated point features.


\paragraph{Curve Feature Propagation (Curve FP)}
Similar to point-based feature propagation~\cite{Qi2017NIPS}, Curve FP is a curve-centric upsampling layer. For each curve $c_j$, this layer propagates features from the low-res polyline $[q_1, ..., q_{L_j}]$ with $L_j$ points to a higher-res polyline $[p_1, ..., p_{H_j}]$ with $H_j$ points where $L_j{<}H_j$. This is achieved using \textit{curve feature interpolation} as described in \cref{subsubsec:cur-ops}. 
Afterwards, the high-res interpolated features are concatenated with skip-linked features from a corresponding curve set abstraction layer and processed by a shared MLP.



\paragraph{Curve Convolution}
The curve convolution layer allows for efficient communication and feature extraction along a curve. This module consists of three sequential \textit{symmetric curve convolutions}, each followed by batch normalization~\cite{Ioffe2015bn} and a leaky ReLU activation.

\vspace{-2mm}
\subsubsection{Curve Operations}
\label{subsubsec:cur-ops}
\vspace{-1mm}
To enable our layers to expressively and efficiently learn on curve clouds, we formulate operations for \textit{sampling}, \textit{grouping}, \textit{feature interpolation}, and \textit{convolutions} along curves.

\paragraph{1D Farthest Point Sampling (FPS)}
FPS is frequently a bottleneck in point cloud architectures and can be costly for large point clouds~\cite{Hu2020CVPR} due to pair-wise distance computations. 
For curves, we alleviate this cost with an approximation of FPS \textit{along each curve} independently in a 1D manner.
This amounts to sampling a subset of points on each curve that are evenly-spaced along the length of that curve (\ie, geodesically).
For a curve $c_j$ with $N_j$ points, we subsample points $[q_1, ..., q_{L_j}]$ with $L_j{<}N_j$ such that all pairs of contiguous points are about $\epsilon$ apart, where $\epsilon$ is a fixed target spacing shared across \textit{all} curves. 
In other words, $d(q_i, q_{i-1}) \approx \epsilon$ for $i=2,\dots,L_j$ where $d$ measures the geodesic distance between two points along the same curve.
Notably, this algorithm has only $O(N)$ complexity and can be parallelized across each curve independently.



\paragraph{Grouping Along Curves}
After sampling, we must group points into local neighborhoods around the subsampled points on each curve. 
We adapt a ball query~\cite{Qi2017NIPS}, which groups together all points within a specified radius from a centroid, to operate along each curve.
For a centroid point $p_i$ belonging to curve $c_j$, we define the local neighborhood of $p_i$ as $\mathcal{N}_i = \{p_k \in c_j \,|\, d(p_i, p_k) < r\}$ where $r$ is a fixed neighborhood radius.
In addition to being computationally faster than a standard 3D ball query grouping, using 1D curve groupings ensures that all neighborhoods lie on a continuous section of scanned surfaces.



\paragraph{Curve Feature Interpolation}
To upsample on curves, we must interpolate features $[g_1, ..., g_{L_j}]$ from a lower-resolution polyline to features $[f_1, ..., f_{H_j}]$ for a higher-resolution one. 
Let $p_h$ be the h$^{th}$ point on the high-res curve, which falls between subsampled low-res points $q_i$ and $q_{i-1}$ with associated features $g_i$ and $g_{i-1}$.
The interpolated high-res feature $f_h$ is simply the distance-weighted interpolation of the two low-res point features (based on their spatial coordinates).



\begin{table*}[t]
\centering
    \centering
    \scalebox{0.75}{
        \setlength{\tabcolsep}{4pt}
        \begin{tabular}{ l | c c c c @{\hspace{1em}}  || @{\hspace{1em}} c | c c c c c c c | c c c}
        \toprule
        & \multicolumn{3}{l}{\textbf{ShapeNet}} & & \multicolumn{8}{l}{\textbf{{KortX}}} & \multicolumn{3}{l}{\textbf{{KortX}}} \\ 
        & \multicolumn{3}{l}{\textit{Per Scan Pattern mIOU ($\uparrow$)}} & & \multicolumn{8}{l}{\textit{Per-Category IOU ($\uparrow$)}} & \multicolumn{3}{l}{\textit{Performance}} \\
        & & & & & \textit{Validation} & \textit{Test} &  & \\
        \textbf{Method} & \rotbox{55}{Mean} & \rotbox{55}{Parallel} & \rotbox{55}{Grid} & \rotbox{55}{Random} & \rotbox{55}{mIOU} & \rotbox{55}{mIOU} & \rotbox{55}{Cap} & \rotbox{55}{Chair} & \rotbox{55}{Earphone} & \rotbox{55}{Knife} & \rotbox{55}{Mug} & \rotbox{55}{Rocket} & \rotbox{55}{Time \scriptsize (ms)} & \rotbox{55}{GPU \scriptsize (GB)} & \rotbox{55}{Param \scriptsize (M)}\\
        \midrule
        PointNet++~\cite{Qi2017NIPS} & 80.1 & 81.8 \scriptsize $\pm$ 0.1 & 80.1 \scriptsize $\pm$ 0.04 & 78.3 \scriptsize $\pm$ 0.7 & 66.9 \scriptsize $\pm$ 1.2 & 71.0 \scriptsize $\pm$ 2.5 & 69.3 & 65.9 & 83.1 & 70.4 & 71.7 & 65.5 & 109 & 1.95 & 1.53 \\
        DGCNN~\cite{Wang2019SIGGRAPHb} & 80.2 & 81.2 \scriptsize $\pm$ 0.2 & 80.9 \scriptsize $\pm$ 0.02 & 78.6 \scriptsize $\pm$ 0.3 & 64.6 \scriptsize $\pm$ 0.8 & 73.3 \scriptsize $\pm$ 0.9 & 73.0 & 76.6 & 81.1 & 79.6 & 58.3 & 71.2 & 143 & 1.45 & 2.18\\
        CurveNet~\cite{Xiang2021ICCV} & 82.8 & \textbf{84.0} \scriptsize $\pm$ 0.2 & \textbf{83.8} \scriptsize $\pm$ 0.4 & 80.6 \scriptsize $\pm$ 0.2 & 68.4 \scriptsize $\pm$ 0.7 & 71.5 \scriptsize $\pm$ 0.4 & 80.9 & \textbf{89.7} & 63.6 & 63.0 & 68.0 & 64.0 & 227 & 1.16 & 5.33 \\
        PointMLP~\cite{Ma2022ICLR} & 80.9 & 82.2 \scriptsize $\pm$ 0.3 & 81.3 \scriptsize $\pm$ 0.3 & 79.3 \scriptsize $\pm$ 0.3 & 72.1 \scriptsize $\pm$ 0.4 & 75.4 \scriptsize $\pm$ 1.3 & 77.2 & 75.3 & 78.8 & 83.1 & 64.6 & 73.6 & 58 & 0.83 & 16.76 \\
        PointNext~\cite{Qian2022PointNeXtRP} & 82.8 & 83.8 \scriptsize $\pm$ 0.1 & 83.1 \scriptsize $\pm$ 0.3 & 81.5 \scriptsize $\pm$ 0.2 & 69.8 \scriptsize $\pm$ 0.6 & 73.7 \scriptsize $\pm$ 0.5 & \textbf{82.2} & 75.9 & 80.9 & 68.3 & 70.8 & 63.9 & 81 & 1.69 & 13.8\\
        \hline  
        MinkowskiNet~\cite{Choy2019CVPR} & 81.1 & 82.6  \scriptsize $\pm$ 0.3 & 82.0  \scriptsize $\pm$ 0.5 & 80.1  \scriptsize $\pm$ 0.5 & 62.9  \scriptsize $\pm$ 1.7 & 60.1 \scriptsize $\pm$ 1.4 & 67.5 & 53.2 & 74.3 & 56.4 & 73.6 & 27.5 & 42 & 0.21 & 36.62\\
        Cylinder3D~\cite{Zhou2020ARXIV} & 79.6 & 81.2 \scriptsize $\pm$ 0.1 & 80.1 \scriptsize $\pm$ 0.0 & 77.6 \scriptsize $\pm$ 0.1 & 58.6 \scriptsize $\pm$ 0.5 & 63.5  \scriptsize $\pm$ 0.2 & 64.8 & 56.9 & 80.8 & 55.1 & 64.8 & 58.7 & 96 & 5.76 & 56.03 \\ 
        SphereFormer~\cite{lai2023spherical} & 79.5 & 80.3 \scriptsize $\pm$ 0.3 & 79.9 \scriptsize $\pm$ 0.4 & 78.3 \scriptsize $\pm$ 0.4  & 67.6 \scriptsize $\pm$ 0.9 & 69.7 \scriptsize $\pm$ 1.3 & 71.5 & 65.1 & 86.0 & 59.6 & 67.2 & \textbf{79.5} & 38 & 0.25 & 32.3\\
        \hline
        CurveCloudNet & \textbf{83.1} & 83.7 \scriptsize $\pm$ 0.2 & 83.6 \scriptsize $\pm$ 0.5 & \textbf{81.9} \scriptsize $\pm$ 0.3 & \textbf{73.0 \scriptsize $\pm$ 0.9} & \textbf{78.9 \scriptsize $\pm$ 1.1} & 69.1 & 87.3 & \textbf{86.7} & \textbf{87.0} & \textbf{74.7} & 67.8 & 77 & 1.01 & 8.74\\
        \bottomrule
        \end{tabular}
    }
    \caption{\textit{Object Segmentation Results.} Class-average mIOU is reported for synthetic ShapeNet dataset (left) and real-world \kortx data (Right). CurveCloudNet achieves the highest accuracy compared to baselines. Performance is on Nvidia RTX 3090 GPU (batch size 16).}
    \label{tab:shapenet-and-kortx}
    \vspace{-4mm}
\end{table*}

\paragraph{Symmetric Curve Convolution}
To process points along curves, we take advantage of expressive convolutions. 
However, it is computationally burdensome to compute neighborhoods on the fly and run convolutions on unordered data \cite{Liu2021TPAMI}.
Instead, we treat each curve as a discrete grid of features that can be convolved similar to a 1D sequence. 
To account for the bi-directionality of curves,  we employ a symmetric convolution and thus produce equivalent results when applied ``forward'' or ``backward'' along the curve. 

In particular, for a curve $c_j = [p_1, p_2, ..., p_{N_j}]$ with associated point features $F_j = [f_1, f_2, ..., f_{N_j}]$, we start by extracting additional features using the the L1 norm of feature gradients along the curve~\cite{Wiersma2022SIGGRAPH}, denoted as $\nabla F_j = \big[ |\nabla f_1|, |\nabla f_2|, ..., |\nabla f_{N_j}| \big ]$. 
Note the norm is necessary to remove directional information.
Concatenating these features together as $[F_j, \nabla F_j] \in \mathbb{R}^{N_j\times D}$ gives a grid on which to perform 1D convolutions.
To respect bi-directionality, symmetric kernels are used for the convolution: for a kernel $W \in \mathbb{R}^{S \times D}$ with size $S$ and $D$ channels, we ensure $W_i = W_{S-i+1}$ for $i = 1, \dots, S$ where $W_i \in \mathbb{R}^D$.




\subsection{Curve Cloud Backbone: \arch} \label{sec:curvecloudnet}
In \cref{fig:curvecloudnet-architecture}, we illustrate CurveCloudNet as designed for segmentation tasks where the output is a single semantic class (one-hot vector) for each point in the input point cloud.
Hence, it follows the U-Net~\cite{Ronneberger2015MICCAI} structure, consisting of a series of downsampling layers followed by upsampling with skip connections.
Although in our experiments (\cref{sec:experiments}) we focus on segmentation, \arch can be adapted to other point cloud perception tasks (see supplement for a classification example).

Our architecture is a mix of curve and point-based layers.
At higher resolutions, curve modules are employed since they are efficient and can capture geometric details when curve sampling is most dense across surfaces in the scene. 
At lower resolutions, point modules are used to propagate information across curves when 1D structure is less apparent. For point operations, we adopt the set abstraction and feature propagation operations from \cite{Qi2017NIPS} as well as the graph convolution from \cite{Wang2019SIGGRAPHb}, and we further improve these point operations following the reportings of recent works \cite{Hu2020CVPR, Ma2022ICLR, Qian2022PointNeXtRP} (see supplementary for further details).
By combining curve and point operations, \arch is an expressive network that maintains the benefits of point cloud backbones while injecting structure and efficiency previously only possible with voxel-based approaches.


\section{Experiments} \label{sec:experiments}

We evaluate CurveCloudNet and a set of competitive baselines on five datasets -- the ShapeNet Part Segmentation dataset~\cite{Chang2015ARXIV}, the KortX Part Segmentation dataset, the Audi Autonomous Driving Dataset (A2D2) \cite{Geyer2020ARXIV}, the nuScenes dataset \cite{Caesar2020CVPR}, and the Semantic KITTI dataset ~\cite{behley2019iccv, Geiger2012CVPR}.
Each dataset exhibits a unique structure and training setup (see ~\cref{tab:summary-results-and-datasets}b). Put together, our evaluation consists of indoor, outdoor, object-centric, scene-centric, sparsely scanned, and densely scanned scenes. Furthermore, each of the five datasets display different point sampling patterns, which we roughly characterize as following ``parallel", ``grid", or ``random" laser motions (see ~\cref{fig:teaser}). 

CurveCloudNet achieves the best or second best performance on \textit{all} datasets, and on average outperforms all previous methods (see \cref{tab:summary-results-and-datasets}a). Furthermore, every other method substantially underperforms \arch on at least one dataset. Notably, \arch outperforms point-based backbones on object-level tasks and is competitive with or better than voxel-based backbones on larger scenes. In \cref{sec:object-segmentation} we evaluate our model on the task of object-level part segmentation on simulated ShapeNet~\cite{Chang2015ARXIV} objects and on a new dataset collected with the \kortx vision system~\cite{summer-robotics}. 
In \cref{sec:audi}, \cref{sec:nuscenes}, and \cref{sec:kitti}, we evaluate semantic segmentation on larger outdoor scenes using the Audi Autonomous Driving Dataset (A2D2) \cite{Geyer2020ARXIV}, the nuScenes dataset \cite{Caesar2020CVPR}, and the Semantic KITTI dataset~\cite{behley2019iccv, Geiger2012CVPR}. 
\cref{sec:ablations} ablates the key components of \arch. In the supplementary, we report additional qualitative results and an experiment on object classification.

\begin{figure}
    \centering
    \includegraphics[width=0.5\textwidth]{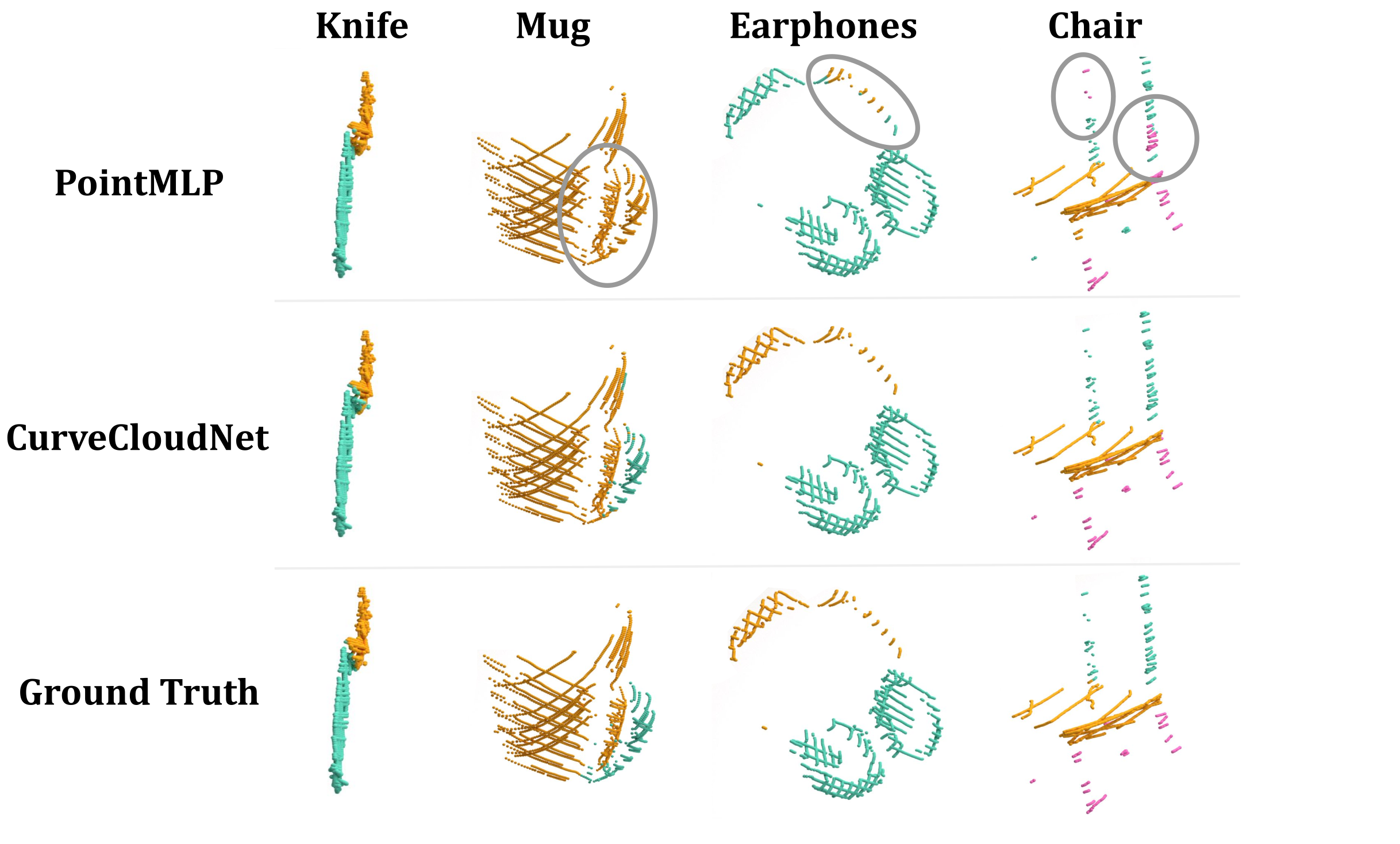}
    \caption{\textit{Qualitative Results on \kortx.} CurveCloudNet successfully segments fine-grained parts by leveraging curve structures.}
    \label{fig:kortx-qual}
    \vspace{-3mm}
\end{figure}
\begin{table*}
\centering
\scalebox{0.82}{
\setlength{\tabcolsep}{4pt}
\begin{tabular}{ l l | c | c c c |  c  c  c  c  c  c  c  c  c  c  c  c  }
\toprule
  & & & \multicolumn{3}{c}{\textbf{Performance} ($\downarrow$)} & \multicolumn{12}{|c}{\textbf{Per-Class mIoU} ($\uparrow$)} \\
  \textbf{Method} & \textbf{Type} & \textbf{mIoU} ($\uparrow$)  & \rotbox{55}{Time \scriptsize (ms)} & \rotbox{55}{GPU \scriptsize (GB)} & \rotbox{55}{Param \scriptsize (M)} & \rotbox{55}{car} & \rotbox{55}{bicycle} & \rotbox{55}{truck} & \rotbox{55}{person} & \rotbox{55}{road} & \rotbox{55}{sidewalk} & \rotbox{55}{obstacle} & \rotbox{55}{building} & \rotbox{55}{nature} & \rotbox{55}{pole} & \rotbox{55}{sign} & \rotbox{55}{signal} \\
  \midrule
  PointNet++~\cite{Qi2017NIPS} & Point & 46.5 & 53 & 0.19 & 1.52 & 62.4 & 9.3 & 55.7 & 3.6 & 90.3 & 58.0 & 12.7 & 79.3 & 82.1 & 19.2 & 36.6 & 48.7 \\
  RandLANet~\cite{Hu2020CVPR} &  & 43.4 & 16 & 0.05 & 1.24 & 60.2 & 4.8 & 46.7 & 7.5 & 91.3 & 57.2 & 14.9 & 78.9 & 80.0 & 16.6 & 27.7 & 34.3 \\
  CurveNet*~\cite{Xiang2021ICCV} & & 4.4 & 385 & 1.77 & 5.52 & 0.0 & 0.0 & 0.0 & 0.0 & 25.4 & 0.0 & 0.0 & 0.0 & 26.9 & 0.0 & 0.0 & 0.0\\
  PointMLP~\cite{Ma2022ICLR} &  & 47.6 & 63 & 0.86 & 16.8 & 65.8 & 9.8 & 54.2 & 15.8 & 92.5 & 63.5 & 14.8 & 81.7 & 82.9 & 18.8 & 34.6 & 36.9 \\ 
  PointNext~\cite{Qian2022PointNeXtRP} & & 45.0 & 34 & 0.33 & 41.6 & 62.6 & 2.6 & 63.1 & 1.0 & 91.1 & 58.7 & 12.8 & 80.1 & 81.8 & 12.5 & 33.8 & 39.3 \\\hline
  MinkowskiNet~\cite{Choy2019CVPR}  & Voxel & 53.8 & 37 & 0.19 & 36.6 & 70.3 & \textbf{13.7} & 77.6 & \textbf{26.8} & 92.8 & 67.5 & 18.0 & 80.8 & 81.9 & 18.0 & 40.5 & 58.1 \\
  Cylinder3D~\cite{Zhou2020ARXIV} & & 53.0 & 61 & 1.18 & 55.8 & 71.1 & 11.6 & 74.8 & 22.1 & 92.5 & 66.1 & 18.3 & 82.6 & 84.2 & 19.1 & 41.6 & 52.0\\ 
  SphereFormer~\cite{lai2023spherical} & & \textbf{55.1} & 54 & 0.37 & 32.3 & \textbf{76.3} & 11.5 & 68.9 & 26.0 & \textbf{93.8} & \textbf{70.1} & 19.6 & \textbf{84.0} & \textbf{86.6} & \textbf{19.8} & \textbf{46.2} & \textbf{58.9} \\
  \hline
  CurveCloudNet & Curve & \underline{54.1} & 75 & 0.27 & 10.2 & 71.9 & 12.9 & \textbf{78.6} & 22.3 & 93.2 & 68.5 & \textbf{19.8} & 83.3 & 85.6 & 17.4 & 44.3 & 51.4\\
  \bottomrule
\end{tabular}}
\vspace{1mm}
\caption{\textit{A2D2 Segmentation Results.} On grid-like \lidar scans, CurveCloudNet outperforms all point-based backbones in mIoU and is competitive with SphereFormer. Performance is on an Nvidia RTX 3090 GPU (batch size 1).}
\label{tab:a2d2-results}
\end{table*}

\subsection{Object Part Segmentation} \label{sec:object-segmentation}

\paragraph{ShapeNet Dataset}
The ShapeNet Part Segmentation Benchmark \cite{Chang2015,Yi2016ToG} contains 16,881 synthetic shape models across 16 different categories and 50 object parts.
To evaluate performance on the laser-based scans that we are interested in, we simulate laser capture using the ShapeNet meshes.
Using a fixed front-facing sensor pose, we raycast a set of linear laser traversals and then sample points on the mesh along each traversal. We consider three types of synthetic laser traversals - \textit{parallel}, \textit{grid}, and \textit{random} - which are depicted on the left side of \cref{fig:teaser} and are further described in the supplementary. We generate one synthetic scan for each ShapeNet mesh, which yields 12139 training point clouds and 1872 validation point clouds.



\paragraph{ShapeNet Results}
ShapeNet results are summarized on the left side of \cref{tab:shapenet-and-kortx}. All methods are trained over three random seeds, and we report the mean and standard deviation of the class-averaged mean intersection-over-union (mIOU) over the runs. For fair comparison, all models are trained for 120 epochs using the same hyperparameters, and the best validation mIOU throughout training is reported.

On average, \arch outperforms all baselines and performs disproportionately well for the ``random" laser traversals. In contrast, SphereFormer exhibits the lowest accuracy of all methods, suggesting that its radial window attention is poorly suited for individual objects. CurveNet and PointNext are the runner ups, showing strong performance on segmenting objects when the scans are captured from the front-facing sensor pose.

\paragraph{\kortx Dataset}
\kortx is a perception software system developed by Summer Robotics~\cite{summer-robotics} that generates and operates on 3D curves sampled from a triangulated system of event sensors and laser scanners. \kortx software supports arbitrary continuous scan patterns, and in practice we scan objects with a  randomly shifted Lissajous trajectory per laser beam. 
Using \kortx, we scan $7$ real-world objects (cap, chair, earphone, knife, mug-1, mug-2, and rocket) multiple times in different poses, collecting 195 point clouds in total.
We will release this dataset upon publication.
We train on scans that are simulated from ShapeNet meshes and evaluate on the \kortx scans as well as the ShapeNet validation split. To best mimic the \kortx data, we simulate \textit{random} laser traversals on each ShapeNet mesh and only train on the six object categories present in the \kortx dataset: \emph{cap, chair, earphone, knife, mug}. We generate five training scans per ShapeNet object, each scanned from a unique \textit{random} sensor pose. This yields a training set of 31,991 point clouds.

\paragraph{KortX Results}
KortX results are summarized on the right side of \cref{tab:shapenet-and-kortx}. The experimental setup is identical to ShapeNet, except we train over four random seeds and for 60 epochs.
CurveCloudNet again outperforms all baselines, showing effective generalization to out-of-domain \kortx test scans. Voxel-based methods continue to underperform their point-based counterparts, suggesting that discretizing the input has a negative effect when point clouds are small. In contrast to the previous ShapeNet evaluation, PointMLP is the second-best method when scans are captured from random sensor poses. 
\cref{fig:kortx-qual} shows that CurveCloudNet better distinguishes fine-grained structures, such as the back of a chair, a mug handle, and an earphone headpiece.

\begin{table*}
\centering
\scalebox{0.77}{
\setlength{\tabcolsep}{3pt}
\begin{tabular}{ l l | c |  c  c  c |  c  c  c  c  c  c  c  c  c  c c  c  c  c c c}
\toprule
  & & & \multicolumn{3}{c}{\textbf{Performance} ($\downarrow$)} & \multicolumn{16}{|c}{\textbf{Per-Class mIoU} ($\uparrow$)} \\
  \textbf{Method} & \textbf{Type} & \textbf{mIoU} ($\uparrow$) & \rotbox{70}{Time \scriptsize (ms)} & \rotbox{70}{GPU \scriptsize (GB)} & \rotbox{70}{Param \scriptsize (M)} & \rotbox{70}{barrier} & \rotbox{70}{bicycle} & \rotbox{70}{bus} & \rotbox{70}{car} & \rotbox{70}{construction} & \rotbox{70}{motorcycle} & \rotbox{70}{pedestrian} & \rotbox{70}{traffic cone} & \rotbox{70}{trailer} & \rotbox{70}{truck} & \rotbox{70}{driveable} & \rotbox{70}{other flat} & \rotbox{70}{sidewalk} & \rotbox{70}{terrain} & \rotbox{70}{manmade} & \rotbox{70}{vegetation} \\ \midrule
  PointNet++~\cite{Qi2017NIPS} & Point & 51.1 &  274 & 0.60 & 1.5 & 60.1 & 6.5 & 58.4 & 66.3 & 16.4 & 20.0 & 50.8 & 12.6 & 31.5 & 42.0 & 94.0 & 60.8 & 63.8 & 69.2 & 82.4 & 82.3\\
  RandLANet~\cite{Hu2020CVPR} &  & 62.9 &  21 & 0.18 & 1.2 & 72.5 & 12.6 & 36.6 & 81.8 & 38.7 & 72.3 & 68.5 & 37.3 & 44.7 & 59.7 & 95.3 & 87.0 & 69.7 & 71.1 & 73.2 & 85.9\\
  CurveNet~\cite{Xiang2021ICCV} & & -- & 
 -- & \textcolor{darkred}{$>$48} & 5.5 & -- & -- & -- & -- & -- & -- & -- & -- & -- & -- & -- & -- & -- & -- & -- & --\\
  PointMLP~\cite{Ma2022ICLR} &  & 67.9 &  164 & 4.94 & 16.8 & 72.3 & 27.8 & 88.2 & 86.3 & 37.2 & 51.0 & 60.7 & 50.6 & 56.4 & 71.1 & 95.7 & 70.6 & 70.9 & 72.0 & 88.8 & 87.2\\
  PointNext~\cite{Qian2022PointNeXtRP} & & 65.0 & 155 & 0.62 & 41.5 & 68.7 & 1.2 & 86.9 & 87.5 & 41.8 & 57.4 & 54.3 & 34.9 & 55.3 & 75.1 & 95.7 & 68.9 & 70.2 & 71.5 & 86.5 & 84.0\\ \hline
  MinkowskiNet~\cite{Choy2019CVPR}  & Voxel & 76.2 & 44 & 0.29 & 36.6 & 75.4 & 43.9 & 91.9 & 93.0 & 49.0 & 84.3 & 78.3 & 64.6 & 65.9 & 85.7 & 96.1 & 71.5 & 67.5 & 74.8 & 86.5 & 84.9 \\
  PolarNet*~\cite{Zhang2020CVPR} & & 71.0 &  - & - & - & 74.7 & 28.2 & 85.3 & 90.9 & 35.1 & 77.5 & 71.3 & 58.8 & 57.4 & 76.1 & 96.5 & 71.1 & 74.7 & 74.0 & 87.3 & 85.7\\
  Cylinder3D*~\cite{Zhou2020ARXIV} & & 76.1 &  80 & 1.57 & 55.9 & 76.4 & 40.3 & 91.2 & 93.8 & 51.3 & 78.0 & 78.9 & 64.9 & 62.1 & 84.4 & 96.8 & 71.6 & 76.4 & 75.4 & 90.5 & 87.4 \\
  SphereFormer*~\cite{lai2023spherical} & & \textbf{79.5} &  
  59 & 0.81 & 32.3 & 78.7 & 46.7 & 95.2 & 93.7 & 54.0 & 88.9 & 81.1 & 68.0 & 74.2 & 86.2 & 97.2 & 74.3 & 76.3 & 75.8 & 91.4 & 89.7\\ \hline
  CurveCloudNet & Curve & \underline{78.0} & 87 & 1.14 & 28.8 & 77.3 & 45.7 & 92.4 & 91.9 & 59.4 & 84.5 & 78.5 & 64.1 & 69.6 & 85.0 & 96.9 & 72.7 & 75.6 & 75.2 & 90.5 & 89.0 \\
  \bottomrule
\end{tabular}}
\vspace{1mm}
\caption{\textit{nuScenes Segmentation Results.} On typical sweeping \lidar scans, CurveCloudNet scales significantly better than other point-based backbones and is competitive with recent work SphereFormer. Performance is on an Nvidia RTX 3090 GPU (batch size 1). * indicates that results are copied from the referenced papers.}
\label{tab:nuscenes-val-results}
\end{table*}

\subsection{A2D2 \lidar Segmentation}\label{sec:audi}

\paragraph{A2D2 Dataset}
The Audi Autonomous Driving Dataset (A2D2) \cite{Geyer2020ARXIV} contains 41,280 frames of outdoor driving scenes captured from 5 overlapping \lidar sensors, creating a unique grid-like scanning pattern (see \cref{fig:overview}).
Each frame is annotated from the front-facing camera with a 38-category semantic label. 
We define a mapping from camera categories to \lidar categories and remove texture-only (\eg sky, lane markers, blurred-area) and very rare categories (\eg animals, tractors, utility vehicles). 
In total, we evaluate on 12 \lidar categories: \emph{car, bicycle, truck, person, road, sidewalk, obstacle, building, nature, pole, sign,} and \emph{traffic signal}. 
Evaluation is performed on annotated LiDAR points in the field of view of the front-facing camera.

\paragraph{Results}
We train CurveCloudNet along with point and voxel-based baselines on the official A2D2 training split~\cite{Geyer2020ARXIV}. 
For fair comparison, all models are trained for 140 epochs using the same hyperparameters, and the best validation mIOU throughout training is reported.
Results are summarized in \cref{tab:a2d2-results}, and \cref{fig:teaser} provides a qualitative example. 
CurveCloudNet scales to outdoor scenes better than point-based backbones, with the runner-up PointMLP showing a 6\% drop in mIOU. CurveCloudNet also outperforms most voxel-based backbones and achieves similar accuracy to state-of-the-art SphereFormer~\cite{lai2023spherical}, even though SphereFormer's radial window attention is tailored for outdoor LiDAR scans.
%


\subsection{nuScenes LiDAR Segmentation} \label{sec:nuscenes}
\paragraph{nuScenes Dataset}
The nuScenes dataset \cite{Caesar2020CVPR} contains 1000 sequences of driving data, each 20 seconds long. 
Each sequence contains 32-beam \lidar data with segmentations annotated at 2Hz. 
We follow the official nuScenes benchmark protocol with 16 semantic categories.

\paragraph{Results}
We train CurveCloudNet and baselines on the official nuScenes training split. 
To ensure fair comparison, we train all models for 100 epochs. 
Results on the nuScenes validation split are shown in \cref{tab:nuscenes-val-results}, and \cref{fig:teaser} includes a qualitative example. 
CurveCloudNet significantly improves upon other point-based networks: PointMLP and PointNext show more than a 10\% drop in mIOU and $\sim 2\times$ increase in latency. We also note that CurveNet exceeds 48GB of GPU memory for a batch size of 1, showcasing its inability to scale to larger scenes.
CurveCloudNet also outperforms all voxel-based methods except the recent SphereFormer.
%



\subsection{KITTI LiDAR Segmentation}\label{sec:kitti}
The Semantic KITTI dataset is made up of 22 sequences of driving data consisting of 23,201 LiDAR scans for training and 20,351 for testing. Each scan is obtained with a dense 64-beam Velodyne LiDAR. We follow the official KITTI protocol in training and validation. To ensure fair comparison, we train all models for 100 epochs. Results on the validation sequence are reported in \cref{tab:kitti-results}, and \cref{fig:teaser} shows a qualitative example. \arch outperforms all point-based and voxel-based methods. Note that we cannot report results for many point-based methods due to excessive training times on the larger KITTI scans ($>20$ days).

\begin{table}
\centering
\setlength{\tabcolsep}{2pt}
\scalebox{0.77}{
\setlength{\tabcolsep}{3pt}
\begin{tabular}{ l l | c |  c  c  c }
\toprule
  & & & \multicolumn{3}{c}{\textbf{Performance} ($\downarrow$)} \\
  \textbf{Method} & \textbf{Type} & \textbf{mIoU} ($\uparrow$) & \rotbox{0}{Time \scriptsize (ms)} & \rotbox{0}{GPU \scriptsize (GB)} & \rotbox{0}{Param \scriptsize (M)} \\
  \midrule
  PointNet++~\cite{Qi2017NIPS} & Point & -- & 2690 & 11.1 & 1.6 \\
  CurveNet~\cite{Xiang2021ICCV} & & -- & -- & \textcolor{darkred}{$>$48} & 5.5 \\
  PointMLP~\cite{Ma2022ICLR} & & 39.5 & 293 & 5.24 & 16.8 \\
  PointNext~\cite{Qian2022PointNeXtRP} & & -- & 1303 & 1.83 & 41.6\\
  \hline
  MinkowskiNet~\cite{Choy2019CVPR} & Voxel & 66.8 & 111 & 0.53 & 36.6 \\
  Cylinder3D*~\cite{Zhou2020ARXIV} & & 68.9 & 233 & 1.62 & 55.9  \\
  SphereFormer*~\cite{lai2023spherical} & & 69.0 & 144 & 3.46 & 32.3\\
  \hline
  CurveCloudNet & Curve & \textbf{69.5} & 155 & 2.75 & 28.8\\
  \bottomrule
\end{tabular}
}
\vspace{-2mm}
\caption{Quantitative results on the KITTI validation split. Performance is on an Nvidia RTX 3090 GPU (batch size 1). * indicates results are copied from the referenced papers.}
\vspace{-1mm}
\label{tab:kitti-results}
\end{table}

\subsection{Ablation Study}\label{sec:ablations}
\begin{table}
\centering
\scalebox{0.84}{
\setlength{\tabcolsep}{3pt}
\begin{tabular}{ c c c | c | c c c}
\toprule
  \multicolumn{3}{c|}{\textbf{Curve Operations}} & & \multicolumn{3}{|c}{\textbf{Performance} ($\downarrow$)} \\
  Grouping & FPS & 1D Conv. & \textbf{mIoU} ($\uparrow$)  & Time \scriptsize (ms) & GPU \scriptsize (GB) & Param \scriptsize (M) \\
  \midrule
   \cmark & \cmark & \cmark & \textbf{54.1} & 75 & 0.27 & 10.3\\
   \xmark & \cmark & \cmark & 53.3 & 99 & 1.03 & 10.3 \\
   \cmark & \xmark & \cmark & 52.4 & 105 & 0.26 & 10.3\\
   \cmark & \cmark & \xmark & 52.0 & \textbf{61} & \textbf{0.20} & 9.9 \\
   \xmark & \xmark & \xmark & 52.6 & 122 & 0.92 & 10.3 \\
  \bottomrule
\end{tabular}}
\vspace{1mm}
\caption{\textit{Ablation Study on A2D2.} Curve operations are ablated and replaced with the standard point-based counterparts.}
\label{tab:a2d2-ablation}
\vspace{-3mm}
\end{table}

\cref{tab:a2d2-ablation} shows an ablation analysis of CurveCloudNet on the A2D2 dataset; the table shows that each of our proposed curve operations is essential to achieve high accuracy and efficiency.
We ablate grouping along curves by instead using the regular radial groupings from PointNet++~\cite{Qi2017NIPS}. This ignores the curve structure and results in decreased accuracy, increased latency, and a significant increase in GPU memory usage. Instead of curve farthest point sampling, we also try regular FPS, which causes a decreased accuracy and increased latency.
Finally, without 1D curve convolutions, we observe a notable decline in accuracy with a marginal improvement in latency and GPU memory. Taken together, our curve operations increase accuracy with roughly half the latency and one third the GPU memory requirements.

%
\section{Discussion and Limitations}


We have described a point cloud processing scheme and backbone, \arch, which introduces curve-level operations to achieve accurate, efficient, and flexible performance on point cloud segmentation. CurveCloudNet outperforms or is competitive with previous methods on the ShapeNet, \kortx, A2D2, nuScenes, and KITTI datasets, and on average achieves the best performance. Put together, \arch is a unified solution to \emph{both} small and large-scale scenes with various scanning patterns.

Nevertheless, \arch has limitations. First, \arch is only designed for laser-scanned data, \ie point clouds with \textit{explicit} curve structure due to 1D laser traversals. We believe a promising future direction is to investigate \textit{virtual curves} that could extend \arch to uniformly sampled point clouds.
Furthermore, we believe that future research can continue to improve curve operations. While our proposed curve operations yield significant improvements, an exciting future direction will be to investigate explicit curve-to-curve communication, curve self-attention and cross-attention, and curve intersections.



\renewcommand{\thesection}{S~\arabic{section}}
\renewcommand{\thetable}{S\arabic{table}}
\renewcommand{\thefigure}{S\arabic{figure}}
\setcounter{table}{0}
\setcounter{figure}{0}
\setcounter{section}{0}

\twocolumn
\newpage
\section{Overview}
In this document, we provide additional method details, dataset details, implementation details, experimental analysis, and qualitative results. In \cref{additional-method}, we concretely outline how we convert a point cloud into a curve cloud and how we implement our 1D farthest point sampling algorithm. In \cref{additional-dataset}, we provide a detailed overview of the Kortx software system and dataset, our ShapeNet simulator, and the A2D2 dataset. In \cref{additional-implementation}, we discuss implementation details of
\arch not covered in the main paper. Finally, in \cref{additional-experiments}, we report results covering GPU memory analysis, object classification, ShapeNet segmentation, and the nuScenes test split.

\section{Additional Method Details} \label{additional-method}

\subsection{Constructing Curve Clouds}
\paragraph{Curve Cloud Conversion}
\begin{figure}[b!]
    \footnotesize
    \centering
    \vspace{-1mm}
    \hrulefill
    \vspace{-1mm}
    \begin{lstlisting}[language=Python]

'''
Inputs: P, T, B, delta
    P: array of size (N, 3) with xyz coordinates
    T: array of size (N,) with timesteps
    B: array of size (N,) with beam IDs
Outputs: curves
    curves: list of arrays, array j is size (N_j, 3)
'''
curves = []
for b in unique(B):
    # filter to a single laser beam's measurments
    beam_P, beam_T = P[beams==b], T[beams==b]

    # order points by laser's traversal
    sequential_ordering = argsort(beam_T)
    beam_P = beam_P[sequential_ordering]

    # split laser's traversal into cont. curves
    edge_lens = norm(beam_P[1:] - beam_P[:-1])
    split_locations = edge_lens > delta

    # convert into polylines
    beam_C = split_seq(beam_P, split_locations)
    curves += beam_C
    \end{lstlisting}
    \hrulefill
    \caption{\textit{Point to Curve Cloud Conversion.} Algorithm (in Python) to convert an input point cloud into a set of polylines.}
    \label{fig:curve-cloud-conversion}
\end{figure}


\paragraph{Constructing Curve Cloud}
We refer the reader to Sec. 3.1 of the main paper for an overview of constructing curve clouds.
As input, we assume that a laser-based 3D sensor outputs a point cloud $P = \{p_1, ..., p_N\}$ where $p_i = [x_i, y_i, z_i] \in \mathbb{R}^3$, an acquisition timestamp $t_i \in \mathbb{R}$ for each point, and an integer laser-beam ID $b_i \in [1, B]$ for each point. We wish to convert the input into a curve cloud $C = \{c_1, ..., c_M\}$, where a curve $c = [p_{i}, ..., p_{i+K}]$ is defined as a sequence of $K$ points where consecutive point pairs are connected by a line segment, \ie, a \textit{polyline}. As outlined in \cref{fig:curve-cloud-conversion}, we first group points by their laser beam ID and sort points based on their acquisition timesteps, resulting in an ordered sequence of points that reflects a single beam's traversal through the scene. Next, for each sequence, we compute the distances between pairs of consecutive points (denoted as polyline ``edge lengths"). Finally, we split the sequence whenever an edge length is greater than a threshold $\delta$, resulting in many variable-length polyline ``curves".  In practice, we parallelize the conversion across all points, and on the large-scale nuScenes dataset, the algorithm runs at 1500Hz.

We select a threshold $\delta$ that reflects the sensor specifications and scanning environment. In particular, the threshold is conservatively set to approximately $10\times$ the \textit{median distance} between consecutively scanned points one meter away from the sensor. On the A2D2 dataset~\cite{Geyer2020ARXIV}, we set $\delta = [0.1, 0.17, 0.1, 0.12, 0.1]$ for the five LiDARs, and on the nuScenes dataset~\cite{Caesar2020CVPR} and KITTI dataset~\cite{behley2019iccv, Geiger2012CVPR} we set $\delta = 0.08$. Additionally, on the A2D2, nuScenes, and KITTI datasets, we scale $\delta$ proportional to the square root of the distance from the sensor, since point samples becomes sparser at greater distance. On the object-level ShapeNet dataset~\cite{Chang2015}, we set $\delta = 0.01$. Experimentally, we observed that \arch is flexible across different $\delta$ values.

\paragraph{Kortx Curve Representation}
The Kortx vision system directly generates and operates on 3D curves sampled from a triangulated system of event-based sensors and laser scanners.
As the detected laser reflection traverses the scene, it produces a frameless 4D data stream that enables low latency, low processing requirements, and high angular resolution. 3D curves are an intrinsic component of the Kortx perception system, and the system directly outputs a curve cloud without the need for additional data processing. 

\subsection{Additional Details on Curve Operations}
\paragraph{1D Farthest Point Sampling}
We refer the reader to Sec. 3.2.2 of the main paper for an overview of our 1D farthest point sampling (FPS) algorithm. The goal of this algorithm is to \textit{efficiently} sample a subset of points on each curve such that consecutive points will be approximately $\epsilon$ apart along the downsampled curve. Concretely, for a curve $c$ with $K$ points, $c = [p_{i}, ..., p_{i+K}]$, we first compute the $K{-}1$ ``edge lengths", $e = [d_{i}, ..., d_{i+K-1}]$, where $d_{i}$ is the distance between consecutive points $(p_{i}, p_{i+1})$. Next, we estimate the geodesic distance along the curve via a cumulative sum operation on the edge lengths: $g = \texttt{CUMSUM}(e)$. Then, we divide the geodesic distances by our desired spacing, $\epsilon$, and take the floor, resulting in $\epsilon$-spaced intervals $I = \texttt{FLOOR}(g / \epsilon)$. We output the first point in each interval, resulting in $L$ $\epsilon$-spaced subsampled points $\{q_1, ..., q_{L}\}$.

For each curve, the computational complexity of the 1D FPS algorithm is $O(K)$. Extending this to all curves, the computational complexity is $O(N)$. When parallelized on a GPU, for each curve, the 1D FPS algorithm has a has \textit{parallel} complexity of $O(\log K)$, where the $\texttt{CUMSUM}$ operation is the parallelization bottleneck (see \textbf{prefix sum} algorithms for more information~\cite{Blelloch1990PrefixSA}). The algorithm trivially parallelizes across curves, leading to a total parallel complexity of $O(\log K)$. In contrast, Euclidean farthest point sampling has a comptational complexity of $O(N^2)$ or $O(N log^2 N)$ (depending on whether a KD-tree is used), and a parallel complexity of $O(L)$. When $L$ is large (\ie we are subsampling a large number of points), Euclidean FPS is a significant performance bottleneck.

\subsection{Additional Details on Point Operations}

\paragraph{Set Abstraction (SA)}
\arch uses a series of set abstraction layers from PointNet++~\cite{Qi2017NIPS}, and we follow previous works to improve the set abstraction layer. First, we perform relative position normalization~\cite{Qian2022PointNeXtRP} -- given a centroid point $p_i$ with local neighborhood points $\mathcal{N}_i$, we center the neighborhood about $p_i$ \textbf{and} we divide by $r$ to normalize the relative positions. Additionally, we opt for the attentive pooling from RandLANet~\cite{Hu2020CVPR} instead of max pooling. 

\paragraph{Graph Convolution}
\arch uses a series of graph convolutions, which are modeled after the edge convolution from DGCNN~\cite{Wang2019SIGGRAPHb}. Unlike DGCNN however, we construct the K-Nearest-Neighbor graph based on 3D point distances instead of feature distances -- this permits more efficient neighborhood construction, irrespective of feature size. Furthermore, we use attentive pooling from RandLANet~\cite{Hu2020CVPR} instead of max pooling.

\section{Additional Dataset Details} \label{additional-dataset}

\subsection{Kortx Perception System and Dataset}
\paragraph{Kortx Perception System}
\kortx is a perception software system developed by Summer Robotics~\cite{summer-robotics}.
It is an active-light, multi-view stereoscopic system using one or more scanning lasers. It can be configured to use two or more event-based vision sensors to build up arbitrary capture volumes.
Event-based vision sensors are used to detect the scanning laser reflection from target surfaces. Event-based sensors are well suited to this setup as their readout electronics are event triggered instead of time triggered.
Furthermore, the Kortx System supports arbitrary continuous scan patterns, allowing a user to create their own patterns and use their own scan hardware.
For more information, please visit the \href{https://www.summerrobotics.ai/}{Summer Robotics Website}.

\begin{figure}
    \centering
    \includegraphics[width=0.5\textwidth]{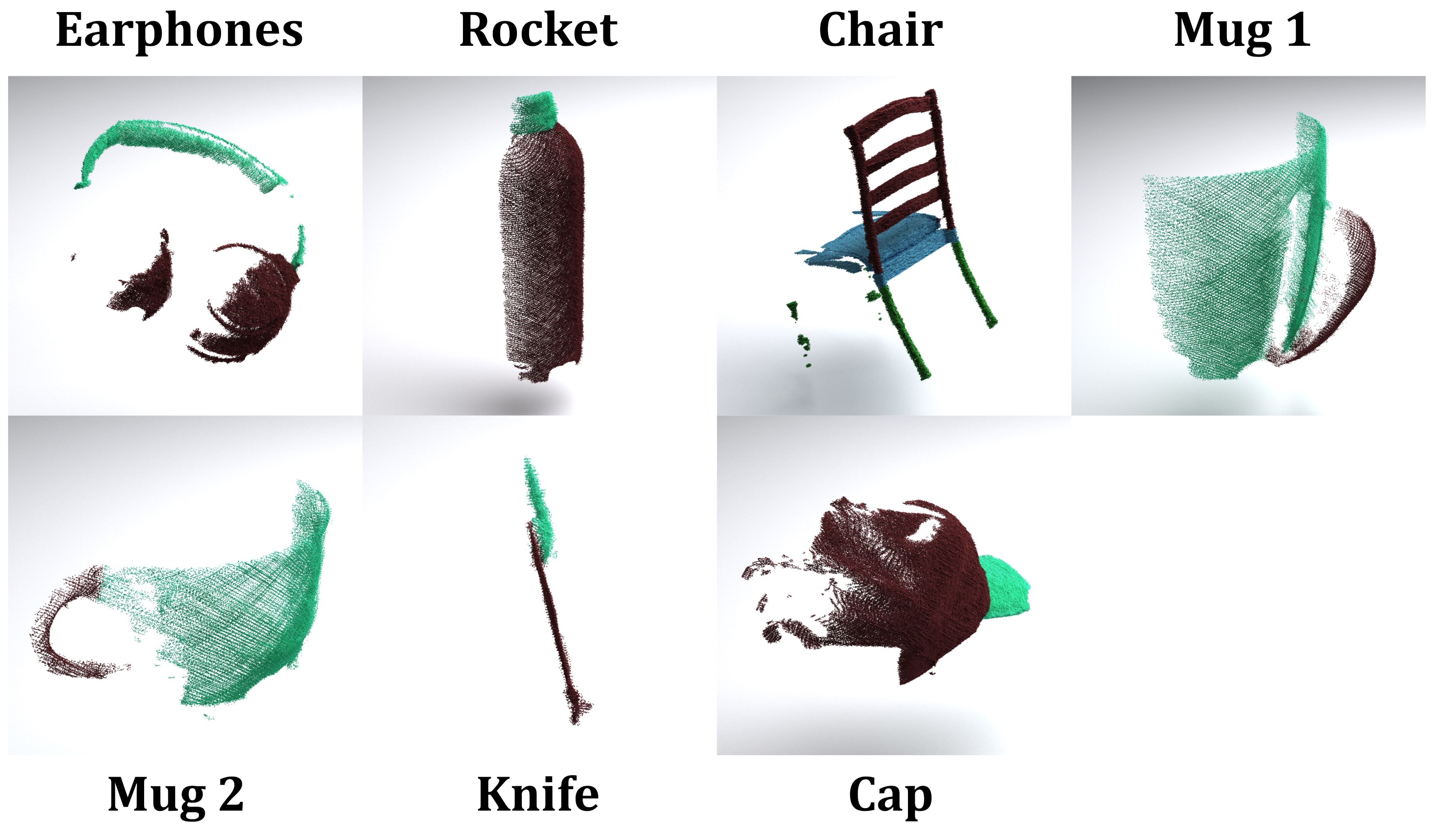}
    \caption{\textit{Kortx Dataset Objects.} Our Kortx dataset contains scans of 7 real-world objects. We visualize one aggregated ``scan" per object from a single viewpoint.}
    \label{fig:kortx-objects-qual}
\end{figure}
\begin{table}
\setlength{\tabcolsep}{3pt}
\centering
\scalebox{0.93}{
    \begin{tabular}{ l c | c c c c c c}
    \toprule
    \textbf{Object} & \textbf{Total} & Cap & Chair & Earphone & Knife & Mug & Rocket\\
    \midrule
    \textbf{Instances} & 7 & 1 & 1 & 1 & 1 & 2 & 1 \\
    \textbf{Scans} & 39 & 6 & 6 & 4 & 6 & 12 & 5 \\
    \textbf{Frames} & 195 & 30 & 30 & 20 & 30 & 60 & 25 \\
    \bottomrule
    \end{tabular}}
    \vspace{1mm}
    \caption{\textit{\kortx Dataset Statistics.} ``Instance" is a unique 3D object. ``Scan" is a dense object scan from a single viewpoint. ``Frame" is a single frame within the 20Hz stream of the dense scan.}
    \label{tab:kortx-dataset}
\end{table}

\paragraph{Kortx Dataset}
Using Kortx, we scanned $7$ real-world objects: \emph{cap, chair, earphone, knife, mug-1, mug-2} and \emph{rocket} (see \cref{fig:kortx-objects-qual}). Each object was scanned multiple times in different poses, resulting in $39$ total scans (summarized in \cref{tab:kortx-dataset}).
Because the Kortx platform provides a continuous event-based 3D scan output (points are sampled every 5µs), we defined a “frame” as a batch of 2048 consecutive point measurements, corresponding to roughly a 20Hz frame rate. Because each frame differs in its dynamic scanning pattern, we evaluate on 5 consecutive frames per scan in our Kortx dataset, hence resulting in 195 point clouds in total.
We manually labeled scanned points with the semantic part categories defined in the ShapeNet Part Segmentation Benchmark \cite{Chang2015}.  Each Kortx scan is mean-centered, however it is \textit{not} aligned into a canonical pose, resulting in an object's orientation depending on the sensor's reference frame. 

\subsection{ShapeNet Simulator} \label{subsec:shapenet-sim}
\begin{figure}[b]
    \centering
    \includegraphics[width=0.48\textwidth]{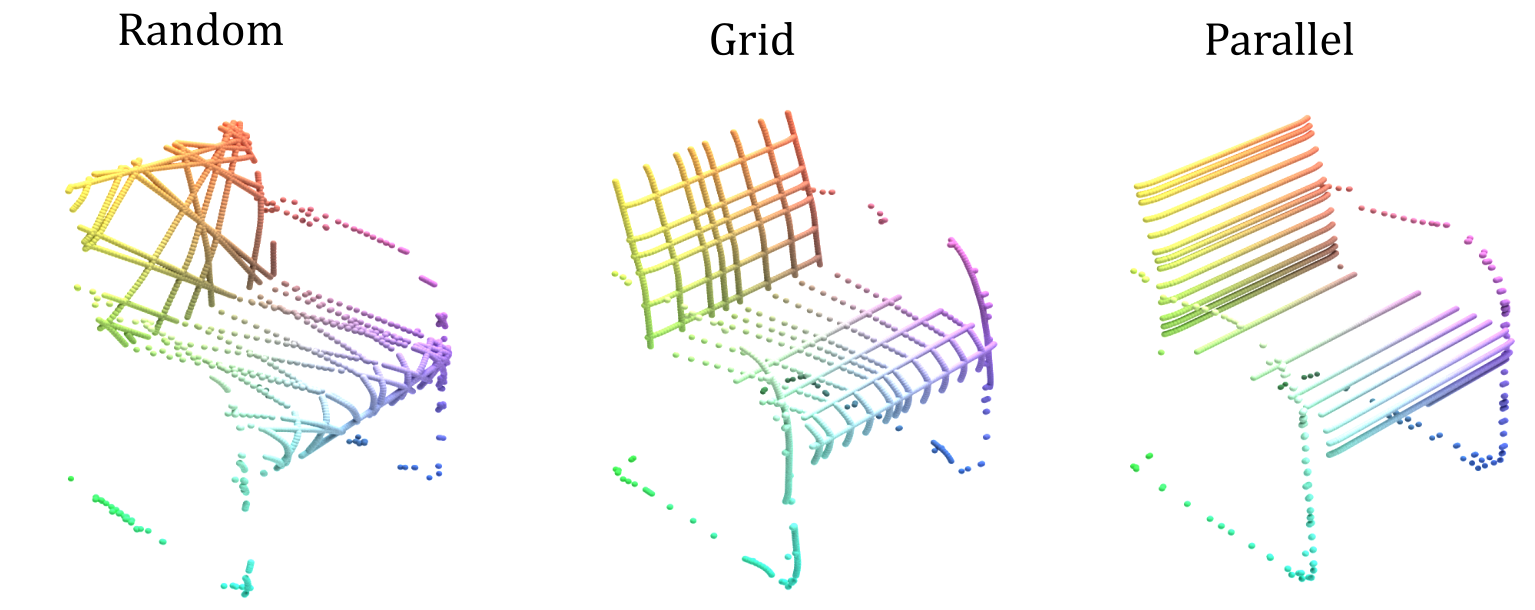}
    \caption{\textit{ShapeNet Simulator.} Our ShapeNet laser-based 3D capture simulator can produce different types of sampling patterns.}
    \label{fig:shapenet-sampling-patterns}
\end{figure}

We simulate laser-based 3D capture on the ShapeNet Dataset \cite{Chang2015}. For each mesh, we randomly sample a sensor pose on the unit sphere and render the mesh's depth values into a $2048\times2048$ image. Next, we sample 2D lines on the depth image that correspond to a laser's traversal. For the \textit{random} sampling pattern used in the Kortx evaluation (see Sec 4.1 of the main paper) and the ShapeNet Classification evaluation (see \cref{subsec:shapenet-class}), we select random linear traversals in the image plane, with each traversal parameterized by a pixel coordinate $(i, j)$ and direction $\theta \in [0, \pi)$. For the \textit{grid} and \textit{parallel} sampling patterns used in our ShapeNet Segmentation evaluation (see Sec 4.1 of the main paper), we sample evenly-spaced vertical and horizontal lines. To reduce descritization artifacts introduced from the rasterization, we query every 6th pixel along each line for the \kortx segmentation task and every 4th pixel for the ShapeNet segmentation and classification tasks. We repeatedly generate synthetic laser traversals until we have sampled 2048 points from the mesh. Fig. 1 of the main paper shows an example of the \textit{random}, \textit{grid}, and \textit{parallel} sampling patterns used in our ShapeNet Segmentation evaluation. \cref{fig:shapenet-sampling-patterns} provides an additional qualitative illustration of the three sampling patterns, but showing 4096 points per scan for greater visual clarity.

\subsection{A2D2 LiDAR Segmentation}
The Audi Autonomous Driving Dataset (A2D2) \cite{Geyer2020ARXIV} contains 41,280 frames of labeled outdoor driving
scenes captured in three cities. The vehicle is equipped with five LiDAR sensors, each mounted on a different part of the vehicle and with a different orientation, resulting in a unique grid-like scanning pattern. The A2D2 data was captured in urban, highway, and rural environments as well as in different weather conditions. At the time of writing, the A2D2 dataset only contains semantic labels for the front-facing camera. Thus, we evaluate on LiDAR observations within the front-facing camera's field of view, and we map camera categories to LiDAR categories. We will release the code detailing the exact mapping.

\subsection{Discussion on 3D Datasets}
As \lidar and other 3D scanning technologies continue to develop, they are being applied to new and diverse applications, including open-world robotics (\ie embedded agents), city planning, agriculture, mining, and more. Additionally, there is an increasing variety of sensors and sensor configurations, spanning hardware that scans at different point densities, different ranges, and with unique (or controllable) scanning patterns. The A2D2 and Kortx datasets are two recent examples of such a trend. We believe an important future direction will be to develop a 3D backbone that is performant in \textit{all} these settings. Furthermore, we believe it is important to understand \textit{which} settings ``break" previous assumptions such as the range-view projection, the birds-eye-view projection, and spherical attention. While \arch is a first step towards this goal, we believe it will be important to capture and compile new 3D datasets, and to evaluate on a greater diversity of environments. 

\section{Additional Implementation Details} \label{additional-implementation}

\subsection{Baselines}

\paragraph{PointNet++ and DGCNN}
We train and evaluate PointNet++~\cite{Qi2017NIPS} and DGCNN~\cite{Wang2019CVPR} using the reproduced implemenations from Pytorch Geometric~\cite{FeyLenssen2019}\footnote{\href{https://github.com/pyg-team/pytorch_geometric/tree/master/examples}{https://github.com/pyg-team/pytorch\_geometric/}}. For PointNet++, we run hyperparameter sweeps to tune the radius and downsampling ratio on each dataset. For DGCNN, we use the authors' reported hyperparameters. 

\paragraph{RandLANet}
We train and evaluate RandLANet~\cite{Hu2020CVPR} using the reproduced implementation from Open3D-ML~\cite{Zhou2018open3d}\footnote{\href{https://github.com/isl-org/Open3D-ML}{https://github.com/isl-org/Open3D-ML}}. We additionally improve the latency by incorporating GPU-implementations for point grouping and sampling from PyTorch3D~\cite{Ravi2020pytorch3d}\footnote{\href{https://github.com/facebookresearch/pytorch3d}{https://github.com/facebookresearch/pytorch3d}}. We use the authors' reported hyperparameters.

\paragraph{CurveNet}
We train and evaluate CurveNet~\cite{Xiang2021ICCV} using the authors' official implementation\footnote{\href{https://github.com/tiangexiang/CurveNet}{https://github.com/tiangexiang/CurveNet}}. We use the authors' reported hyperparameters for all datasets.

\paragraph{PointMLP}
We train and evaluate PointMLP~\cite{Ma2022ICLR} using the authors' official implementation\footnote{\href{https://github.com/ma-xu/pointMLP-pytorch}{https://github.com/ma-xu/pointMLP-pytorch}}. We use the reported hyperparameters for all datasets.

\paragraph{PointNext}
We train and evaluate PointNext~\cite{Qian2022PointNeXtRP} using the authors' official implementation\footnote{\href{https://github.com/guochengqian/PointNeXt}{https://github.com/guochengqian/PointNeXt}}. As outlined by the authors, we use PointNext-Small for the ShapeNet and KortX datasets. On the A2D2, nuScenes, and KITTI datasets, we use the larger PointNext-XL. Because the authors indicate the importance of the network ``radius", we additionally performed a hyperparameter sweep to find the best radius of 0.05 for the A2D2, nuScenes, and KITTI datasets.  

\paragraph{MinkowskiNet}
We train and evaluate MinkowskiNet~\cite{Choy2019CVPR} using the authors' official implementation\footnote{\href{https://github.com/NVIDIA/MinkowskiEngine}{https://github.com/NVIDIA/MinkowskiEngine}}. We use the larger MinkUNet-34A for all experiments. We use an initial voxel size of $0.05$ on outdoor datasets and $0.015$ on object-level datasets. 

\paragraph{Cylinder3D}
We train and evaluate Cylinder3D~\cite{Zhou2020ARXIV} using the authors' official implementation\footnote{\href{https://github.com/xinge008/Cylinder3D}{https://github.com/xinge008/Cylinder3D}}. We use the reported hyperparameters on the nuScenes and KITTI datasets. On the A2D2 dataset, we set the cylindrical voxel grid to cover a $\pm 31 ^{\circ}$ forward-facing azimuth with a maximum radius of 80 meters and a height covering $[-5, 20]$ meters; we define the initial grid to have 360 radial partitions, 120 angular partitions, and 120 height partitions. On the ShapeNet and KortX datasets, we set the voxel grid to cover all $360^{\circ}$ with a radius of $1.0$ and height of $1.0$; to address latency and memory constraints, we define the initial grid to have 96 radial partitions, 96 angular partitions, and 96 height partitions.

\paragraph{SphereFormer}
We train and evaluate SphereFormer~\cite{lai2023spherical} using the authors' official implementation\footnote{\href{https://github.com/dvlab-research/SphereFormer}{https://github.com/dvlab-research/SphereFormer}}. We use the reported hyperparameters for the nuScenes and KITTI datasets, and we use the reported nuScenes hyperparameters for the A2D2 dataset. For the KortX and ShapeNet datasets, we also use the reported hyperparameters, and we reduce the voxel size from $0.1$ to $0.015$ to account for the dataset's smaller 3D scale. On the KortX and ShapeNet datasets, we additionally ran a sweep on different voxel sizes and spherical window sizes, but observed limited differences.

\subsection{Training Strategy}
We train \arch and baselines on segmentation tasks with a standard cross-entropy loss. Following previous works, we also supplement the loss with a Lovasz loss \cite{Berman2018CVPR, Zhou2020ARXIV} for the nuScenes, A2D2, and KITTI datasets. At training, we apply random scaling and translation augmentations, as well as random flips on the nuScenes, A2D2, and KITTI datasets. Importantly, we use an \textbf{\textit{identical}} training strategy for \arch and each baseline. We experimentally observe convergence in all models' validation accuracies by the end of training.


\paragraph{Object Part Segmentation}
We train CurveCloudNet and all baselines for 60 epochs in the KortX experiment and 120 epochs in the ShapeNet experiment with the Adam optimizer \cite{Kingma2015ICLR}, a learning rate of $1e^{-4}$, batch momentum decay of $0.97$, and exponential learning rate decay of $0.97$. For
all models, except for Cylinder3D, we use a batch size of 24. For Cylinder3D, we use a batch size of 12 because 24 exceeds our GPU memory capacity. 

\paragraph{A2D2 LiDAR Segmentation}
We train CurveCloudNet and all baselines for 140 epochs with the Adam optimizer, a batch size of 7, a learning rate of $1e^{-3}$, and an exponential learning rate decay of $0.97$. 

\paragraph{nuScenes and KITTI LiDAR Segmentation}
We train CurveCloudNet and all baselines for 100 epochs with the Adam optimizer, a batch size of 4 on nuScenes and 2 on KITTI, a learning rate of $1e^{-3}$, and an exponential learning rate decay of $0.97$. At test time, we follow previous works \cite{Zhou2020ARXIV,Liu2021TPAMI} and average model predictions over axis-flipping and scaling augmentations.


\section{Additional Experiments} \label{additional-experiments}

\begin{table*}
\centering
\scalebox{0.82}{
\setlength{\tabcolsep}{4pt}
\begin{tabular}{ l | c |  c  c  c  c  c  c  c  c  c  c  c  c  }
\toprule
  & & \multicolumn{12}{|c}{\textbf{Per-Class mIoU} ($\uparrow$)} \\
  \textbf{Method} & \textbf{mIoU} ($\uparrow$) & \rotbox{55}{car} & \rotbox{55}{bicycle} & \rotbox{55}{truck} & \rotbox{55}{person} & \rotbox{55}{road} & \rotbox{55}{sidewalk} & \rotbox{55}{obstacle} & \rotbox{55}{building} & \rotbox{55}{nature} & \rotbox{55}{pole} & \rotbox{55}{sign} & \rotbox{55}{signal} \\
  \midrule
  MinkowskiNet~\cite{Choy2019CVPR} & 42.7 & 54.7 & 3.1 & 60.0 & 7.4 & 87.3 & 54.8 & 9.7 & 71.8 & 76.7 & 14.3 & 31.5 & 42.4 \\
  SphereFormer~\cite{lai2023spherical} & 40.5 & 52.7 & 1.1 & 62.4 & 3.8 & 85.9 & 53.5 & 10.9 & 70.5 & 76.3 & 9.6 & 29.1 & 29.9 \\
  \hline
  CurveCloudNet & \textbf{44.8} & 58.4 & 2.2 & 59.8 & 6.7 & 89.9 & 58.5 & 11.1 & 77.8 & 83.3 & 13.1 & 35.6 & 42.2\\
  \bottomrule
\end{tabular}}
\vspace{1mm}
\caption{\textit{Translated A2D2 Results.} When A2D2 scans are randomly translated, CurveCloudNet significantly outperforms SphereFormer, suggesting that SphereFormer relies on a dataset exhibiting a highly consistent global structure.}
\label{tab:translated-a2d2-results}
\end{table*}

\subsection{Translated A2D2}
\paragraph{Overview}
In Sec. 4.2 of the main paper, we reported that SphereFormer outperforms \arch by $+1.0\%$ mIOU on the A2D2 dataset. In this section, we show that, on the same data, SphereFormer underperforms \arch when the scene does not exhibit consistent and aligned global structure. 

We apply a simple translation augmentation to the A2D2 training and validation data -- for each scan, we offset all points by a translation sampled from a uniform Gaussian with $\mu=0$ and $\sigma=20$. Note that this removes the \textit{global} alignment of point clouds, but completely preserves all \textit{local} structure. In the real world, this setup could occur in topography or mapping, \ie when a large region is scanned but only one area is of interest (which could be anywhere in the scan). We train and evaluate all models with an identical setup to the original A2D2 experiment.

\paragraph{Results}
We summarize results on the translated A2D2 experiment in \cref{tab:translated-a2d2-results}. Without a consistent global alignment of the scene layout, all methods perform worse. However, \arch is less effected and outperforms SphereFormer by over 4$\%$. This further suggests that SphereFormer's radial window is tailored for outdoor driving scenes and cannot be flexibly applied to environments with weaker global structure. In contrast, \arch can successfully leverage local structures to reason in more diverse environments. 

\begin{table*}
\setlength{\tabcolsep}{4pt}
\centering
\scalebox{0.93}{
    \begin{tabular}{ l | c c c | c c c }
    \toprule
    & \multicolumn{3}{c}{\textbf{{Accuracy}}} & \multicolumn{3}{c}{\textbf{{ Performance}}} \\
    \textbf{Method} & Class ($\uparrow$) & Instance ($\uparrow$) & F1 ($\uparrow$) & Time \scriptsize (ms) ($\downarrow$) & GPU \scriptsize (GB) ($\downarrow$) & Param \scriptsize (M) \\
    \midrule
    PointNet++~\cite{Qi2017NIPS} & 95.3 \scriptsize $\pm$ 0.7 & 99.0 \scriptsize $\pm$ 0.05 & 95.5 \scriptsize $\pm$ 0.6 & 51 & 0.91 & 1.6 \\
    DGCNN~\cite{Wang2019SIGGRAPHb} & 93.7 \scriptsize $\pm$ 0.5 & 98.9 \scriptsize $\pm$ 0.03 & 93.6 \scriptsize $\pm$ 0.5 & 73 & 0.78 & 0.6 \\
    PointMLP~\cite{Ma2022ICLR} & 94.8 \scriptsize $\pm$ 1.3 & 99.2 \scriptsize $\pm$ 0.05 & 95.3 \scriptsize $\pm$ 1.0 & 54 & 0.76 & 13.2 \\ \hline
    CurveCloudNet & \textbf{96.3 \scriptsize $\pm$ 0.4} & \textbf{99.3 \scriptsize $\pm$ 0.04} & \textbf{96.0 \scriptsize $\pm$ 0.5} & 37 & 0.66 & 10.3\\
    \bottomrule
    \end{tabular}}
    \vspace{1mm}
    \caption{\textit{Object Classification Results.} Mean class-averaged accuracy (Class), instance-averaged accuracy (Instance), and class-averaged F1 score (F1) are reported for the ShapeNet data. CurveCloudNet outperforms baselines on all metrics. Performance is on an Nvidia RTX 3090 GPU (batch size 16).}
    \label{tab:shapenet-classification}
\end{table*}

\subsection{ShapeNet Classification}\label{subsec:shapenet-class}
In addition to semantic segmentation tasks, we also evaluate CurveCloudNet's performance in shape classification.

\paragraph{ShapeNet Classification Dataset}
We use the ShapeNet Part Segmentation Benchmark \cite{Chang2015,Yi2016ToG} as described in Sec 4.1 of the main paper. In the classification setting, the network is tasked with classifying a point cloud into one of the 16 object categories. Using our ShapeNet laser-scanner simulator (see \cref{subsec:shapenet-sim}), we generate a single synthetic ``scan" for each ShapeNet mesh from a \emph{fixed} sensor viewpoint, resulting in scanned objects sharing a canonical orientation. Following the official training and validation splits \cite{Yi2016ToG}, this yields 12139 training point clouds and 1872 validation point clouds.  For this experiment, we consider the \textit{random} curve sampling pattern (see \cref{subsec:shapenet-sim}).

\paragraph{Setup}
We train CurveCloudNet and several baselines on the simulated ShapeNet training set. Similar to the settings used for the segmentation task, all models are trained for 120 epochs with the Adam optimizer, a batch size of 24, a learning rate of $3e^{-4}$, batch momentum decay of 0.97, and exponential learning rate decay of 0.97. We record the best validation class-averaged accuracy, instance-averaged accuracy, and class-averaged F1 score that is achieved during training. We report means and standard deviations across 3 runs.

\paragraph{Results}
Results are summarized in \cref{tab:shapenet-classification}. CurveCloudNet outperforms the baselines on all three metrics. Additionally, CurveCloudNet exhibits improved latency and lower GPU memory compared to PointNet++, DGCNN, and PointMLP. 

\subsection{Additional nuScenes Results}
\begin{table*}
\centering
\scalebox{0.77}{
\setlength{\tabcolsep}{3pt}
\begin{tabular}{ l l |  c  c |  c  c  c  c  c  c  c  c  c  c c  c  c  c c c}
\toprule
  & & & & \multicolumn{16}{|c}{\textbf{Per-Class mIoU} ($\uparrow$)} \\
  \textbf{Method} & \textbf{Type} & \textbf{mIoU} ($\uparrow$) & \textbf{fwIoU} ($\uparrow$) & \rotbox{70}{barrier} & \rotbox{70}{bicycle} & \rotbox{70}{bus} & \rotbox{70}{car} & \rotbox{70}{construction} & \rotbox{70}{motorcycle} & \rotbox{70}{pedestrian} & \rotbox{70}{traffic cone} & \rotbox{70}{trailer} & \rotbox{70}{truck} & \rotbox{70}{driveable} & \rotbox{70}{other flat} & \rotbox{70}{sidewalk} & \rotbox{70}{terrain} & \rotbox{70}{manmade} & \rotbox{70}{vegetation} \\ \midrule
  PolarNet~\cite{Zhang2020CVPR} & Voxel & 69.4 & 87.4 & 72.2 & 16.8 & 77.0 & 86.5 & 55.1 & 69.7 & 64.8 & 54.1 & 69.7 & 63.5 & 96.6 & 67.1 & 77.7 & 72.1 & 87.1 & 84.5\\
  Cylinder3D~\cite{Zhou2020ARXIV} & & 77.2 & 89.9 & 82.8 & 29.8 & 84.3 & 89.4 & 63.0 & 79.3 & 77.2 & 73.4 & 84.6 & 69.1 & 97.7 & 70.2 & 80.3 & 75.5 & 90.4 & 87.6 \\
  SPVNAS~\cite{Liu2021TPAMI} & & 77.4 & 89.7 & 80.0 & 30.0 & 91.9 & 90.8 & 64.7 & 79.0 & 75.6 & 70.9 & 81.0 & 74.6 & 97.4 & 69.2 & 80.0 & 76.1 & 89.3 & 87.1\\
  AF$^2$-S3Net~\cite{Cheng2021AF2S3NetAF} & & 78.3 & 88.5 & 78.9 & 52.2 & 89.9 & 84.2 & 77.4 & 74.3 & 77.3 & 72.0 & 83.9 & 73.8 & 97.1 & 66.5 & 77.5 & 74.0 & 87.7 & 86.8 \\
  SphereFormer~\cite{lai2023spherical} & & \textbf{81.9} & \textbf{91.7} & 83.3 & 39.2 & 94.7 & 92.5 & 77.5 & 84.2 & 84.4 & 79.1 & 88.4 & 78.3 & 97.9 & 69.0 & 81.5 & 77.2 & 93.4 & 90.2\\
  \hline
  CurveCloudNet & Curve & \underline{78.5} & \underline{90.4} & 81.7 & 38.4 & 93.8 & 90.3 & 70.1 & 77.3 & 75.1 & 69.5 & 84.9 & 74.1 & 97.5 & 67.8 & 79.7 & 75.7 & 91.7 & 88.7 \\
  \bottomrule
\end{tabular}}
\vspace{1mm}
\caption{\textit{nuScenes Test Split.} \arch demonstrates improved or competitive performance with top-performing baselines.}
\label{tab:nuscenes-test-results}
\end{table*}

\begin{figure*}
    \centering
    \includegraphics[width=1.0\textwidth]{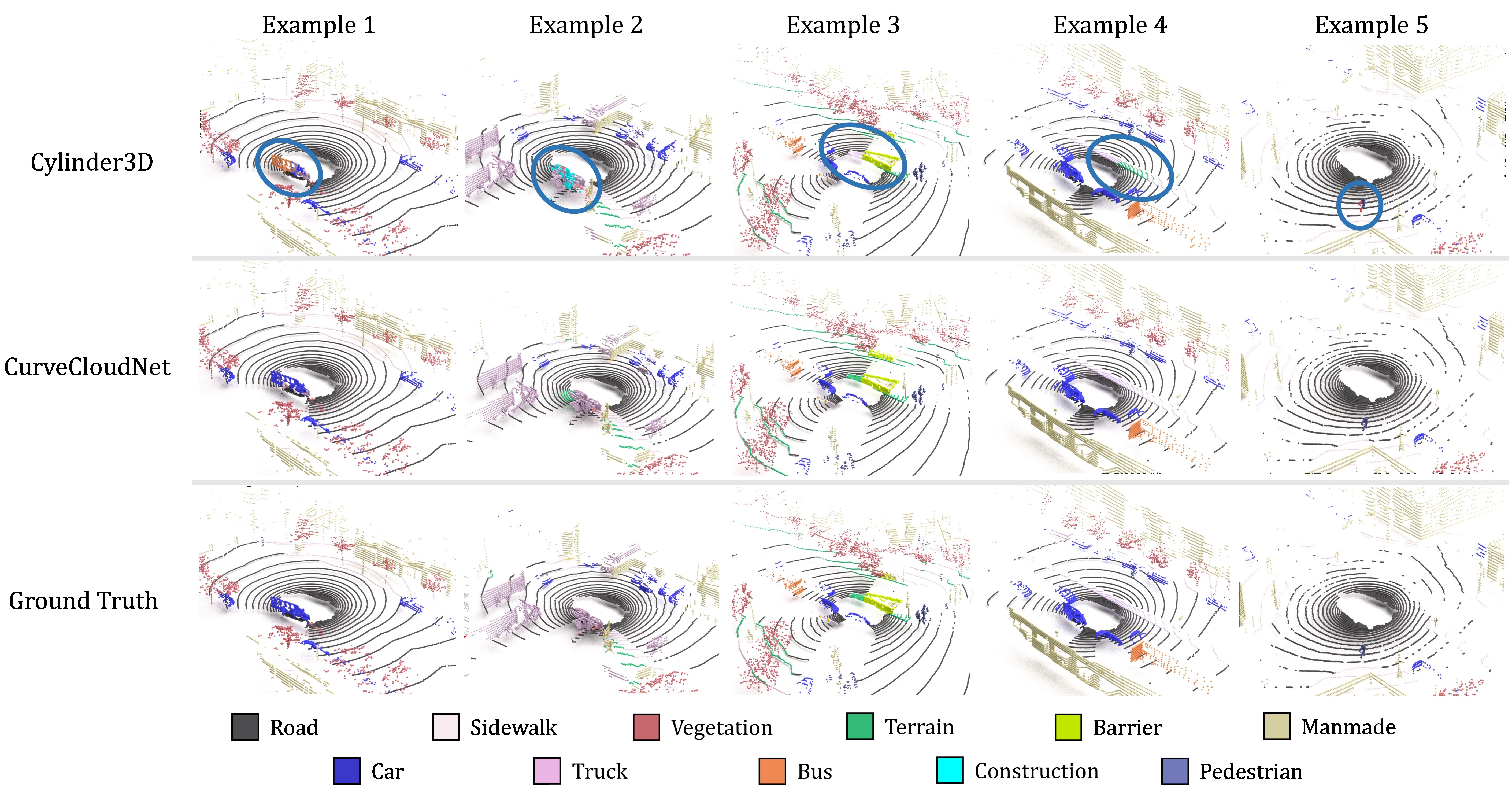}
    \caption{\textit{nuScenes Segmentation.} We visualize semantic segmentation predictions of CurveCloudNet and Cylinder3D on the nuScenes validation split. We highlight regions where Cylinder3D errors and \arch predicts correctly.}
    \label{fig:nuscenes-qual}
\end{figure*}

We provide qualitative results on the nuScenes validation split in \cref{fig:nuscenes-qual}. For the corresponding evaluation on the validation split, see Sec. 4.3 of the main paper.

\subsubsection{nuScenes Test Split}
We evaluate our model on the test split of the nuScenes LiDAR segmentation task, and compare to top-performing baselines from the academic literature.
We summarize our results in \cref{tab:nuscenes-test-results}. \arch outperforms many sparse-voxel methods in both class-averaged and frequency-weighted mIOU, such as Cylinder3D~\cite{Zhou2020ARXIV}, SPVNAS~\cite{Liu2021TPAMI}, and AF$^2$-S3Net~\cite{Cheng2021AF2S3NetAF}.



\subsection{Additional KortX Results}
\begin{figure*}
    \centering
    \includegraphics[width=1.0\textwidth]{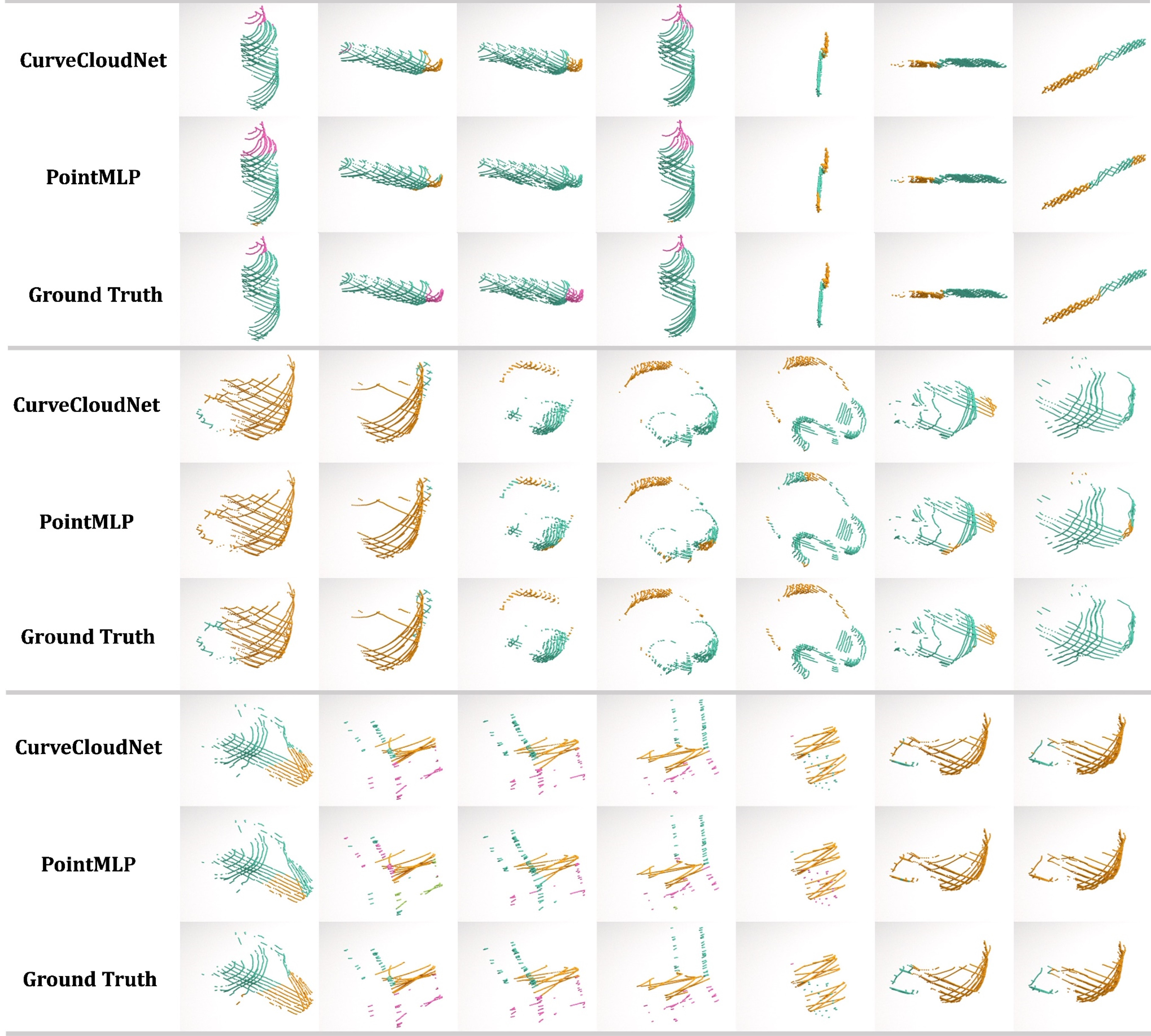}
    \caption{\textit{Kortx Dataset Part Segmentation.} We visualize segmentation predictions of CurveCloudNet and PointMLP on the KortX dataset.}
    \label{fig:kortx-qual-fullpage}
\end{figure*}

We provide additional qualitative results on the KortX dataset in \cref{fig:kortx-qual-fullpage}. We observe that \arch distinguishes finegrained structures, such as the legs and back of the chair, the handle of the mug, the boundary where the nose of the rocket begins, and the brim of the cap.

\subsection{Additional A2D2 Results}
\begin{figure*}
    \centering
    \includegraphics[width=1.0\textwidth]{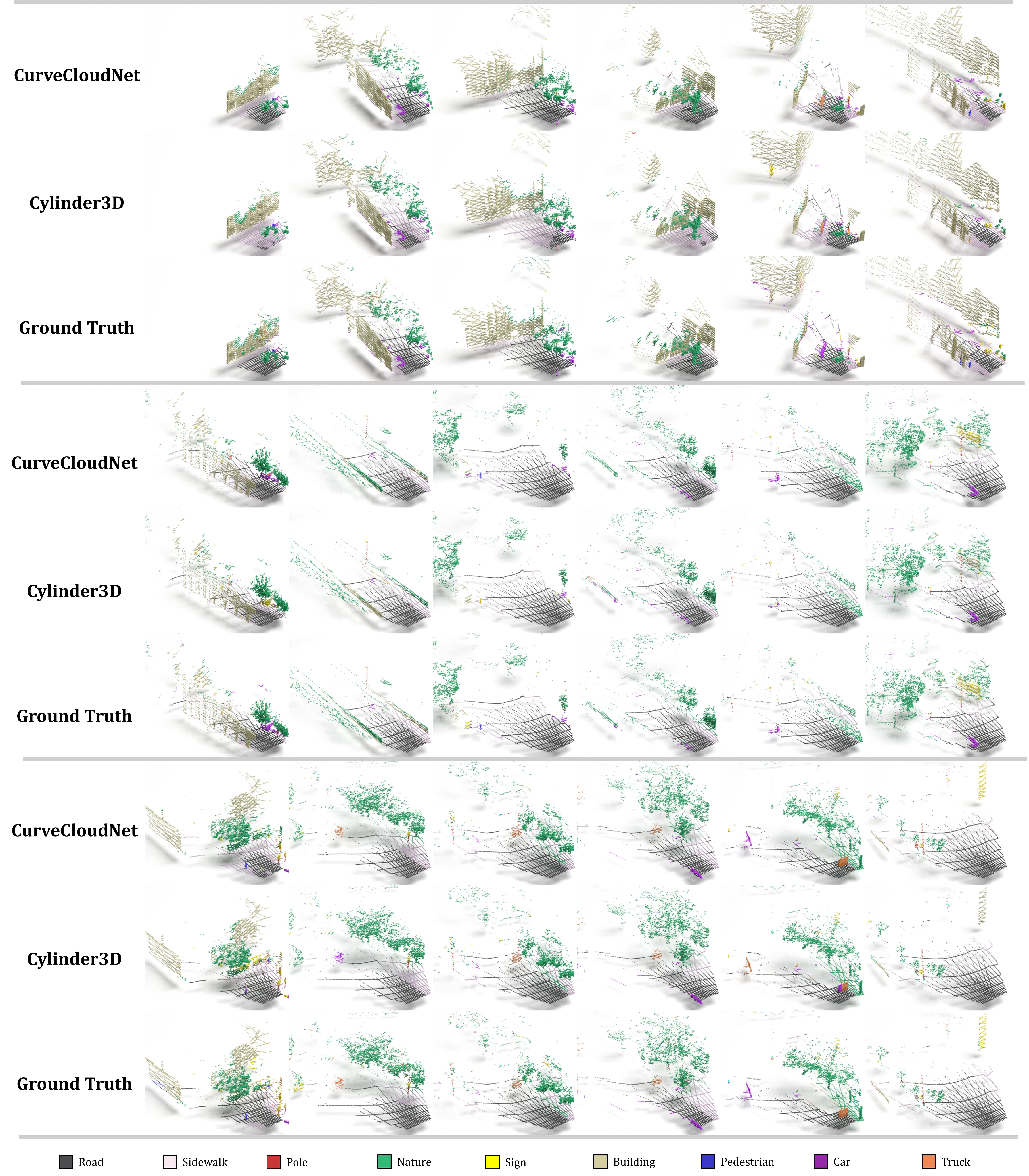}
    \caption{\textit{A2D2 Dataset Segmentation.} We visualize segmentation predictions of CurveCloudNet and Cylinder3D on the A2D2 dataset.}
    \label{fig:audi-qual-fullpage}
\end{figure*}

We provide additional qualitative results on the A2D2 dataset in \cref{fig:audi-qual-fullpage}. In many examples, \arch distinguishes the sidewalk and the road much better than Cylinder3D. 
Furthermore, in contrast to \arch, examples show that Cylinder3D can fail to detect pedestrians, swaps the ``truck" and ``car" categories, and swaps the ``building" and ``sign" categories.

\newpage


{
    \small
    \bibliographystyle{ieeenat_fullname}   \bibliography{bibliography_long,bibliography,bibliography_custom}
}

\end{document}